%% file: main.tex
\documentclass{article}

\usepackage[final]{neurips_2022}

\usepackage[utf8]{inputenc} 
\usepackage[T1]{fontenc}    
\usepackage{xr-hyper}
\usepackage[hidelinks]{hyperref}       
\usepackage{url}            
\usepackage{booktabs}       
\usepackage{amsfonts}       
\usepackage{nicefrac}       
\usepackage{microtype}      
\usepackage{xcolor}         
\usepackage{placeins}
\usepackage{graphicx}
\usepackage{multirow}
\usepackage{lineno}
\usepackage{amsmath}
\usepackage{dsfont}
\usepackage{colortbl}
\usepackage{comment}
\usepackage{enumitem}
\usepackage{cleveref}
\usepackage{wrapfig}
\usepackage[normalem]{ulem}

\excludecomment{separateappendix}
\includecomment{concatenatedappendix}
\excludecomment{checklist}

\begin{separateappendix}
	\newcommand{\aref}[1]{\ref*{#1}}
\end{separateappendix}
\begin{concatenatedappendix}
	\newcommand{\aref}[1]{\ref{#1}}
\end{concatenatedappendix}

\title{Self-Supervised Learning Through Efference Copies}

\author{%
	Franz Scherr$^{1}$\textbf{*}\\
	Huawei Technologies\\
	{\footnotesize \texttt{franz.scherr@huawei.com}}\\
	\And
	Qinghai Guo$^2$\\
	Huawei Technologies \\
	{\footnotesize \texttt{guoqinghai@huawei.com}} \\
	\And
	Timoleon Moraitis$^{1}$\textbf{*}\\
	Huawei Technologies\\
	{\footnotesize \texttt{timoleon.moraitis@huawei.com}} \\
}
\begin{document}

\footnotetext[1]{Huawei Zurich Research Center, Switzerland, $^2$Huawei ACS Lab, Shenzhen, China, \textbf{*}Corresponding author}

\maketitle

\begin{abstract}
Self-supervised learning (SSL) methods aim to exploit the abundance of unlabelled data for machine learning (ML), however the underlying principles are often method-specific.
An SSL framework derived from biological first principles of embodied learning could unify the various SSL methods, help elucidate learning in the brain, and possibly improve ML.
SSL commonly transforms each training datapoint into a pair of views, uses the knowledge of this pairing as a positive (i.e.\ non-contrastive) self-supervisory sign, and potentially opposes it to unrelated, (i.e.\ contrastive) negative examples.
Here, we show that this type of self-supervision is an incomplete implementation of a concept from neuroscience, the Efference Copy (EC).
Specifically, the brain also transforms the environment through efference, i.e.\ motor commands, however it sends to itself an EC of the full commands, i.e.\ more than a mere SSL sign. In addition, its action representations are likely egocentric.
From such a principled foundation we formally recover and extend SSL methods such as SimCLR, BYOL, and ReLIC under a common theoretical framework, i.e.\ Self-supervision Through Efference Copies (S-TEC). Empirically, S-TEC restructures meaningfully the within- and between-class representations. This manifests as improvement in recent strong SSL baselines in image classification, segmentation, object detection, and in audio. These results hypothesize a testable positive influence from the brain's motor outputs onto its sensory representations.
\end{abstract}

\section{Introduction}
\begin{figure}
	\centering
	\includegraphics[width=13.9cm]{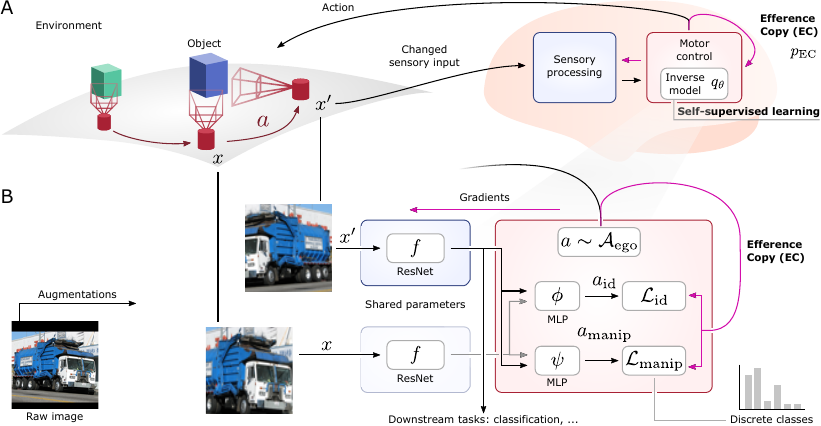}
	\caption{\textbf{Efference copy (EC). A}) Sensory-motor system: Efferent actions $a$ (change in focus, manipulations, etc.) yield changed sensory input $x'$. The internal copy of the motor command, i.e.\ the EC, we propose, may be used as a self-supervisory signal for learning an inverse model $q_\theta$. \textbf{B}) Abstract ML setting: The action space comprises switching and augmenting images. Sensory processing and inverse models are implemented as neural networks, trained through EC's feedback.}
	\label{fig:principle}
\end{figure}
Deep Learning (DL) has drawn inspiration from Neuroscience and also offers models for understanding aspects of the brain~\citep{richards2019deep}.
DL has been extremely successful, largely owing to labelled big datasets. However, such labelling is a costly procedure carried out by human supervisors. 
Fully unsupervised ML techniques do exist, however they rarely reach the performance of supervised learning \citep{moraitis2022softhebb, journe2022hebbian}. On the other hand, recently, a category of algorithms that are self-supervised has emerged.
In self-supervised learning (SSL), the model itself generates the supervisory signal, so that human supervision is not needed, and then uses that signal for supervised learning.
Recent SSL algorithms generate the supervisory signal by using pairs of inputs where it is known whether they are of the same or of a different instance, therefore self-generating positive or negative labels. 
Examples can be associated as being positive, e.g.\ based on their temporal proximity, if the input is in a sequential domain~\citep{oord2018representation}. 
Advanced SSL algorithms generate themselves the positive pairs of inputs, by augmenting the training dataset~\citep{he2020momentum, chen2020simple, grill2020bootstrap, caron2020unsupervised, mitrovic2021representation}. 
Further improving these algorithms even has the potential to outperform supervised learning~\citep{tomasev2022pushing}, because more information exists in the comparison of complete input pairs than in individual human-labelled examples.
Therefore, improving SSL consists in devising representation-learning methods that better capture that information. However, conceptual frameworks that unify these principles of existing SSL methods, and guide towards new improved ones, are scarce~\citep{balestriero2022contrastive}.

\section{Efference copies in the central nervous system}
\label{sec:ec}
The operation of the biological central nervous system (CNS), which appears to learn mostly without external supervision, may provide such a framework. Conversely, ML simulations within such a framework may also generate testable hypotheses for biological SSL.
In the present study we take this abstract hope and formulate it as a concrete link from SSL to a specific mechanism in the CNS.
We begin by observing first, that the CNS of vertebrate animals is believed to have evolved with the main purpose of performing sensory-motor control and learning, and second, that the data manipulations that augment the training examples in ML implementations of SSL can be viewed as motor actions.
The search for analogies then can focus on looking for possible self-supervisory signals within biological motor control and learning.

A particularly well-suited and well-established signal in the sensory-motor system is that of the Efference Copy (EC)~\citep{vonhelmholtz1867handbuch, mcnamee2019internal} or Corollary Discharge~\citep{sperry1950neural}.
Namely, it has been shown that when a component of the CNS addresses the body's muscles with an efferent, i.e.\ outgoing, motor command or action, often it also sends a copy to the CNS itself, see Fig.~\ref{fig:principle}A.
ECs have multiple functions and abundant supporting evidence~\citep{kennedy2014temporal, mcnamee2019internal, kilteni2020efference, latash2021efference}.
For example, certain motor commands responsible for the locomotion of frogs are generated in the spinal cord, but are copied to the brainstem, which is responsible for motor control of the eye~\citep{vonuckermann2013spinal}.
The body-movement-related disturbances to the visual field are then predicted and appropriately counteracted by eye movements that stabilize the frog's gaze.
Therefore, one function of EC is to coordinate different motor controllers of the body.
Another function of EC is to focus sensory processing on externally-generated and unpredicted stimuli by cancelling predictable sensations of self-generated actions.
E.g., humans cannot tickle themselves effectively, because by using its ECs the CNS predicts the sensory consequence of its own action, and cancels it before it is perceived~\citep{blakemore1998central}.
The role of ECs in humans is actually broader and very central to motor control.
Specifically, the control of bodily movements involves forward internal models that the brain maintains, i.e.\ models that predict the sensory inputs that result from each motor command~\citep{kawato1999internal, mcnamee2019internal}.
These forward models rely on access to motor commands to generate their predictions, and that access is provided by ECs. Importantly, motor control also involves inverse models, which map representations of targeted movement sensations to their possible actions~\citep{rizzolatti1998organization,kawato1999internal} (Fig.~\ref{fig:principle}A).
In addition to motor control, ECs also underlie motor \textit{learning}~\citep{witney1999predictive, troyer2000associational, diedrichsen2003anticipatory, engert2013fish, brownstone2015spinal}.
For example, when learning an inverse model, the structures that calculate errors must access the efference.

Given the pervasive role of ECs in sensory processing, motor control, and motor learning, we hypothesize that ECs could play a key role in the learning of sensory representations too, and that it does so through the learning of inverse models.
More specifically, we hypothesize that, if the EC acts as a self-supervisory learning signal, then it improves the sensory learning process, e.g.\ improving the later classification of input examples.
Rather than physiological experiments, or biologically detailed simulations, we will test the hypothesis in an abstract ML setting.
Nevertheless, we will use mechanisms that do have plausible biophysical implementations.
In addition, our model could improve ML methods by providing more of the information content of paired input datapoints to SSL.
That is because ECs can be rich and diverse signals, i.e.\ they can provide the full description of the actions that generate input pairs, and can do so for varied types of actions.

\section{S-TEC: Self-supervision Through Efference Copies}
\label{sec:stec}
\begin{figure}
	\centering
	\includegraphics[width=13.9cm]{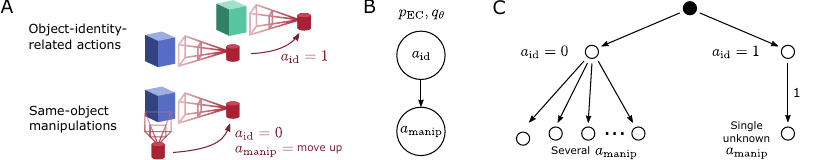}
	\caption{\textbf{Actions. A}) Categories: New objects can be brought into focus $a_\mathrm{id}=1$ (e.g.\ saccades). Else ($a_\mathrm{id}=0$), the same object can be manipulated by $a_\mathrm{manip}$ (e.g.\ moving). \textbf{B}) Dependency structure as a graphical model. \textbf{C}) Corresponding decision tree if $a_\mathrm{manip}$ is further assumed to be discrete.}
	\label{fig:actions}
\end{figure}
\subsection{Definitions and key principles}
\label{sec:definitions}
Our basic assumption is that \textit{\textbf{(a) an EC is available}}, i.e.\ a copy of the efferent motor commands, or actions.
The essence of our strategy is to use this EC as a target label to \textit{\textbf{(b) learn an inverse model}}, mapping sensory inputs to the motor outputs that caused the changed inputs in the first place~\citep{kawato1999internal}.
We assume \textit{\textbf{(c) a hierarchical model}}, e.g.\ a multilayer neural network.
We conjecture that a model that improves on this motor-oriented task, will also improve its intermediate sensory representations as a direct consequence, which are then useful to a wider variety of sensory tasks.
In our experiments, we use the representation for classification.
In order to concretize the model, let $x \in \mathcal{D}$ generally denote sensory inputs.
Furthermore, we define the motor commands as actions $a \in \mathcal{A}_\text{ego}$ that result in transformed inputs $x'=T(x,a)$, denoting with $T$ the transformation function, see also Fig.~\ref{fig:principle}A.
In the following, we will simply write the EC as a probability distribution $p_\mathrm{EC}(a|x,x')$ to indicate the distribution of values it will assume given the sensory inputs are $x$ before an action was taken, and are $x'$ thereafter.
We utilize this as a ground truth that the inverse model needs to predict.
More formally, we denote the to-be-learned inverse sensory-motor mapping by $q_\theta(a|x,x')$ with free parameters $\theta$ that surmise synaptic weights.
Learning then is the minimization of the discrepancy between the ground truth $p_\mathrm{EC}$ and our model $q_\theta$:
\begin{align}
	\min_\theta \mathop{\mathds{E}}_{\substack{x,\tilde a\in \mathcal{D} \times \mathcal{A}_\text{ego}\\ x'=T(x, \tilde a)}} \Big[\, \underbrace{D_\mathrm{KL}( p_\mathrm{EC}(a|x, x') ; q_\theta(a|x, x') )}_{=: \mathcal{L} ~\text{(Loss)}} \,\Big]~.\label{eq:principle}
\end{align}
In the above formulation, we denote with $D_\mathrm{KL}$ the Kullback-Leibler divergence, and introduce the loss function $\mathcal{L}$ that will be helpful later.
The broad concept given in Eq.~\eqref{eq:principle} is so far agnostic to the specific types of actions and sensory inputs.
To render the matter more concrete, and to align it with the examples for sensory modalities in Section~\ref{sec:ec}, we will focus on the visual sensory domain.
This also facilitates the validation of our approach by ML experiments on contemporary datasets and architectures, see Section~\ref{sec:results}.
We assume that only one type of sensory object is observed with each sensory input.
We denote the set of possible actions as $\mathcal{A}_\text{ego}$.
To account for the fact that the model concerns sensory-motor control in the physical world, \textbf{\textit{(d) actions must account for two types of sensory transformations}} (see Fig.~\ref{fig:actions}A), namely:
\begin{itemize}[topsep=0em,itemsep=0.5em,partopsep=0em, parsep=0em]
	\item \emph{\textbf{(d1) Object-identity-related actions} $a_\mathrm{id}$.} This category of action switches between sensed objects, e.g.\ by a saccade of the eyes, bringing entirely new objects into focus, or not.
In the context of standard vision datasets that are comprised of static images, we simply exchange the currently viewed image with a randomly sampled new one. The two types of actions in this category, i.e.\ switching or not, are $a_\mathrm{id} = 1$ or $a_\mathrm{id}=0$ respectively.
	\item \emph{\textbf{(d2) Same-object manipulations} $a_\mathrm{manip}$.} 
	This category is identity-preserving, i.e.\ maintaining the sensed object but the observer actively manipulates it or its view,
	e.g.: turning the object, or moving to a closer vantage point.
	With static images, this kind of transformation is naturally formed by commonly used image augmentation operations.
E.g.\ spatial transformations that crop an image with random size and random aspect ratio can simulate the movement to a different point at a closer distance whereas mirroring the image horizontally corresponds well to rotating a symmetric 3D object, see Fig.~\ref{fig:principle}B.
	We denote an action that transforms one augmented view $x$ into the other augmented view $x'$ by $a_\mathrm{manip}$.
\end{itemize}

The object-identity-related actions are of two types $a_\mathrm{id}\in\{0,1\}$. Therefore, this part of the action representation is categorical. Based on this, \textbf{\textit{(e) we model the entire action representation as categorical}}, i.e.\ including $a_\mathrm{manip}$. This is to follow the biological evidence that the brain maintains uniform principles throughout its organization, e.g.\ throughout the cortex~\citep{douglas1989canonical}, including motor areas~\citep{bastos2012canonical}.
Moreover, there is significant evidence that this uniform organization does specifically have a categorical structure, were different actions are represented by different clusters of neurons~\citep{graziano2016ethological}. 
Importantly, this allows learning the associated inverse model by means of a classification task, as will be introduced later.

As the overall action $a$ is composed by two parts, i.e.\ $a=(a_\mathrm{id}, a_\mathrm{manip})$, we can summarize this categorical structure as a graphical model and decision tree, shown in Fig.~\ref{fig:actions}B and C.
Hence, given the same object continues to be in focus, i.e.\ $a_\mathrm{id}=0$, then there exist several options for the object-manipulating action $a_\mathrm{manip}$.
In the other case, where focus is switched to a different object, i.e.\ $a_\mathrm{id}=1$, there is no value of the object-manipulation action $a_\mathrm{manip}$ that relates $x$ and $x'$.
To formally represent this in the decision tree, we assign all probability to some unknown $a_\mathrm{manip}$ in that case.

So far we have not described how each class of action $a_\mathrm{manip}$, and therefore its copy EC, is parametrized by the motor controller. Based on the fact that EC conveys to the observer the action that himself is taking, it is appropriate to \textit{\textbf{(f) use an egocentric representation of actions}} instead of aligning the actions with an allocentric reference point, i.e.\ with the environment.
To do so, notice that the spatial transformations introduced in (d2) are affine, thus can be represented with their associated transformation matrices.
This logical parametrization allows us to conveniently compute the egocentric action that is needed to turn $x$ into $x'$: By multiplication of the transformation matrix that gave rise to one view from the original with the inverted transformation matrix that gave rise to the other.
This highlights a difference to the allocentric representation of actions that was chosen in other work~\citep{lee2021improving}, where transformations were aligned to the original, allocentric reference frame (i.e.\ differences of scales rather than their quotient as it would emerge here).

\subsection{Formalism}
\label{sec:formalism}
Through the preceding dependency structure (Fig.~\ref{fig:actions}B, C), the inverse model naturally decomposes into two more specific inverse models, where one is attributed to object-identity-related actions, and the other to the same-object manipulations, to which we simply refer to as ``\textbf{identity-related inverse model}'' $q_\theta(a_\mathrm{id}|x,x')$ and ``\textbf{manipulation-related inverse model}'' $q_\theta(a_\mathrm{manip}|a_\mathrm{id}, x, x')$ respectively.
Therefore we have that $q_\theta(a|x,x') = q_\theta(a_\mathrm{id}| x, x') q_\theta(a_\mathrm{manip}|a_\mathrm{id}, x, x')$.
Applying the same also for the ground truth $p_\text{EC}$ enables us to split the loss function into separate parts ${\mathcal{L}=\mathcal{L}_\mathrm{id} + \mathcal{L}_\mathrm{manip}}$ that reflect learning of the identity-related inverse model and learning of the manipulation-related inverse model correspondingly.
More precisely, the loss dedicated to the identity-related inverse model is given by ${\mathcal{L}_\mathrm{id} = D_\mathrm{KL}( p_\mathrm{EC}(a_\mathrm{id}|x, x') ; q_\theta(a_\mathrm{id}|x, x') )}$, while similarly, the loss dedicated to the manipulation-related inverse model is given by ${\mathcal{L}_\mathrm{manip} = D_\mathrm{KL}( p_\mathrm{EC}(a_\mathrm{manip}|a_\mathrm{id},x, x') ; q_\theta(a_\mathrm{manip}|a_\mathrm{id},x, x') )}$, see also Appendix~\aref{sec:derivation} for details.

In practice, we also include regularization losses $\mathcal{L}_\mathrm{reg}$, see Appendix~\aref{sec:optimization}, and weight the relative importance of the loss terms by hyperparameters $\lambda$.
Therefore, the loss that we consider is given by:
\begin{align}
	\mathcal{L} &= \mathcal{L}_\mathrm{id} + \lambda_\mathrm{manip} \mathcal{L}_\mathrm{manip} + \lambda_\mathrm{reg} \mathcal{L}_\mathrm{reg}~.\label{eq:general_loss}
\end{align}
\paragraph{Instantiation of the inverse models.}
The specific formulation of $q_\theta(a|x,x')$ determines the loss function that will be optimized.
We consider several options for the formulation of $q_\theta(a_\mathrm{id}|x,x')$ that allow us to recover various contemporary approaches for SSL as we discuss in Results Section~\ref{sec:recover}.
On the other hand, for the manipulation-related inverse model, we opted for a categorical action representation that clusters similar actions as advocated in key principle (e).
We implemented this by subdividing the support of each single component $a_{k,\mathrm{manip}}$ of $a_\mathrm{manip}$ (consisting of 6 components for the affine transformation) into a number of $K$ discrete bins. We indicate these discretized versions of the real actions with a hat $\widehat{\cdot}$ and define the probability of $a_{k,\mathrm{manip}}$ being in bin $b$ as:
\begin{align}
	q_\theta(\widehat a_{k,\mathrm{manip}}=b|a_\mathrm{id}=0,x,x') = \frac{\exp \left(\psi_{k, b}(f(x), f(x')) \right)}{\sum_j \exp \left(\psi_{k, j}(f(x), f(x')) \right)}~,\label{eq:aug}
\end{align}
where we have introduced a feature extractor $f$ (ResNets in our case, see Fig.~\ref{fig:principle}) and the functions $\psi_{k,j}$ (for which we used MLPs) to express the model's belief that $a_{k,\mathrm{manip}}$ assumes a value in discrete bin $j$.
Note that the functions $f$ and $\psi$ both are learnable, but the dependence on $\theta$ is omitted for brevity.
We refer to Fig.~\ref{fig:hp} for an ablation study on alternative instantiations of the manipulation-related inverse model.

\section{Results}
\label{sec:results}

\subsection{Recovering contrastive \& non-contrastive SSL from the identity-related inverse model}
\label{sec:recover}
Depending on the specific instantiation of the identity-related inverse model, we recover several common approaches for SSL using the concept of ECs.
In particular, we show that based on the choice of the learned $q_\theta(a_\mathrm{id}|x,x')$, we can recover from the identity-related loss either contrastive losses (i.e.\ instance discrimination) such as employed in SimCLR~\citep{chen2020simple}, ReLIC~\citep{mitrovic2021representation} or ReLICv2~\citep{tomasev2022pushing}, or non-contrastive approaches such as BYOL~\citep{grill2020bootstrap} (see below, and Appendices~\aref{sec:ntxent_recovery},~\aref{sec:byol_recovery}, and~\aref{sec:relic_recovery}).

First, we consider here as an example the identity-related inverse model that gives rise to the contrastive loss of SimCLR~\citep{chen2020simple}.
We define this inverse model's probability of no identity-switch, i.e. $a_\mathrm{id}=0$, in the common way used for the positive view in contrastive learning~\citep{chen2020simple}, for which we adopt the notation provided by~\citet{mitrovic2021representation}:
\begin{align}
	q_\theta(a_\mathrm{id}=0|x,x') = \frac{\exp\left( \phi(f(x), f(x')) / \tau \right)}{ \sum_{x_n \in \{x'\} \cup\,C}\exp\left( \phi(f(x), f(x_n)) / \tau \right)}~.\label{eq:sw}
\end{align}
Here, $\phi$ is a function that computes a similarity between the features produced by $f$, i.e.\ it compares the intermediate sensory representations.
We defined it as a scalar product between projected features: $\phi(h, h'):= \sum_i g_i(h) g_i(h')$, whereas $g$ is a multi-layer perceptron (MLP), following typical choices in the literature (see also Appendix~\aref{sec:architectures}).
The scalar $\tau$ is a temperature hyperparameter, and the set $C$ is composed of additional candidate inputs to which $x$ is compared to (through the denominator).
Note that $g$ is learnable also, thus depending on $\theta$.

We assume that the EC is a perfect copy of $a$, hence $p_\text{EC}(a|x,x')$ assigns all probability to the true action $a$ that was applied.
In doing so, we obtain an upper bound of the objective~\eqref{eq:general_loss} that we use for our S-TEC experiments (see Appendix~\aref{sec:derivation} for the derivation).
Its associated component dedicated to the identity-related inverse model is the typical contrastive learning objective:
\begin{align}
	\mathcal{L}\leq&-\log \frac{\exp(\phi(f(x), f(x'')) / \tau)}{\sum_{x_n \in \{x'\} \cup\,C}\exp\left( \phi(f(x), f(x_n)) / \tau \right)} \nonumber \\
	&\quad - \lambda_\mathrm{manip} \sum_k \log \frac{\exp \left(\psi_{k, j''}(f(x), f(x'')) \right)}{\sum_j \exp \left(\psi_{k, j}(f(x), f(x'')) \right)} + \lambda_\mathrm{reg} \mathcal{L}_\mathrm{reg}~,\label{eq:objective}
\end{align}
where we introduced $x''$ to always represent an input that is related to $x$ through $a_\mathrm{id} = 0$.
We use $x''$ also for the second term that concerns the manipulation-related inverse model, to reflect that the loss is only applied in that condition.
For additional convenience we write $j''$ to refer to the bin in which the value of $a_{k,\mathrm{manip}}$, relating $x$ and $x''$, falls.

By integrating a confidence-estimate into the identity-related inverse model, we obtain and extend SSL methods such as ReLIC and ReLICv2.
Furthermore, by defining the identity-related inverse model as a normal distribution, we obtain non-contrastive losses such as that of BYOL. For these derivations, see Appendices~\aref{sec:byol_recovery} and~\aref{sec:relic_recovery}. Interestingly, the loss for VICReg~\citep{bardes2022vicreg} is of the same type as derived from S-TEC's principles for a normally-distributed identity-related inverse model.

\subsection{Experimental evaluation}
\begin{table}[h]
	\caption{Accuracies obtained with linear classification (mean and std over 5 independent runs).}
	\label{tab:results}
	\centering
	\begin{tabular}{llccc}
		\toprule
		Architecture & Method     & CIFAR-10     & CIFAR-100 & STL-10 \\
		\midrule
		\multirow{6}{*}{ResNet-18}
		& SimCLR {\scriptsize repr. \citep{chen2020simple}} & 91.5 {\scriptsize $\pm 0.1$}  & 65.3 {\scriptsize $\pm 0.3$} & \textcolor{black}{\textbf{91.5}} {\scriptsize $\pm 0.4$}  \\
		& ReLIC {\scriptsize repr. \citep{mitrovic2021representation}}& \textcolor{black}{91.5 {\scriptsize $\pm 0.2$}} & 65.6 {\scriptsize $\pm 0.3$} & \textcolor{black}{\textbf{91.4} {\scriptsize $\pm 0.2$}}  \\
		& MoCo v2 {\scriptsize repr. \citep{chen2020improved}} & 90.9 {\scriptsize $\pm 0.2$} & 65.7 {\scriptsize $\pm 0.4$} & \textcolor{black}{88.6 {\scriptsize $\pm 0.6$}} \\ 
		& BYOL {\scriptsize repr. \citep{grill2020bootstrap}} & {92.0} {\scriptsize $\pm 0.2$} & 66.4 {\scriptsize $\pm 0.4$} & \textcolor{black}{91.1 {\scriptsize $\pm 0.2$}} \\
		& S-TEC {\scriptsize (ours)}& {92.0} {\scriptsize $\pm 0.2$} & 66.6 {\scriptsize $\pm 0.3$} & \textcolor{black}{\textbf{91.6}} {\scriptsize $\pm 0.2$} \\ 
		& S-TEC$^*${\scriptsize (ours)}& \textbf{92.6} {\scriptsize $\pm 0.1$} & \textbf{67.6} {\scriptsize $\pm 0.4$} & \textcolor{black}{\textbf{91.4} {\scriptsize $\pm 0.2$}} \\
		\midrule
		\multirow{4}{*}{ResNet-50} & SimCLR {\scriptsize repr. \citep{chen2020simple}} & 93.2 {\scriptsize $\pm 0.2$} & \textcolor{black}{70.8} {\scriptsize $\pm 0.2$}  & \textcolor{black}{94.0} {\scriptsize $\pm 0.1$} \\
		& ReLIC {\scriptsize repr. \citep{mitrovic2021representation}} & \textcolor{black}{93.2 {\scriptsize $\pm 0.2$}} & \textcolor{black}{70.8 {\scriptsize $\pm 0.2$}}  & \textcolor{black}{93.8 {\scriptsize $\pm 0.1$}} \\
		& S-TEC {\scriptsize (ours)} & \textbf{93.9} {\scriptsize $\pm 0.2$} & \textcolor{black}{71.9} {\scriptsize $\pm 0.4$} & \textcolor{black}{\textbf{94.3}} {\scriptsize $\pm 0.2$} \\
		& S-TEC$^*${\scriptsize (ours)} & \textcolor{black}{\textbf{94.0} {\scriptsize $\pm 0.1$}} & \textcolor{black}{\textbf{72.4} {\scriptsize $\pm 0.1$}} & \textcolor{black}{93.9 {\scriptsize $\pm 0.2$}}  \\
		\bottomrule
	\end{tabular}
\end{table}

\begin{figure}[b]
	\includegraphics[width=13.9cm]{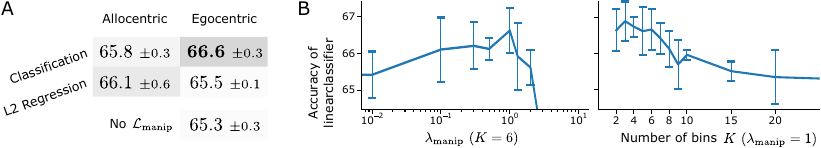}
	\caption{\textbf{Ablation study on CIFAR-100 with ResNet-18. A)} Instantiation of the manipulation-related inverse model and action representation. \textbf{B)} Hyperparameter sweep. For any setting we report mean and 95\% confidence interval based on $\geq$5 independent runs.}
	\label{fig:hp}
\end{figure}

In the preceding Sections we have derived our framework that connects the concept of ECs to current methods for SSL.
Here, we present the results of our experimental evaluations that aim to assess the quality of the representations that can be learned with our approach.
For this purpose, we considered various image datasets, including CIFAR-10/100~\citep{krizhevsky2009learning}, STL-10~\citep{coates2011stl10} as well as the ImageNet ILSVRC-2012 dataset~\citep{imagenet}, and compare S-TEC against several other SSL algorithms, such as SimCLR~\citep{chen2020simple}, MoCo v2~\citep{chen2020improved}, BYOL~\citep{grill2020bootstrap} and ReLIC~\citep{mitrovic2021representation}.
We follow the same procedure for all experiments, where we first performed SSL and subsequently determine the class prediction accuracy of a linear classifier that is trained on the emergent representations.
During SSL the same data augmentation methods as in~\citet{chen2020simple} are applied throughout, i.e.\ also colour augmentations, but these were not considered for training an inverse model with S-TEC, see also Appendix~\aref{sec:input_transformations} for details to the augmentations applied.
We adopted the ResNet v1 framework~\citep{he2016deep}, and used specifically ResNet-18 and ResNet-50 architectures as the feature extractor $f$, while $\phi$ and $\psi$ were generally implemented as multi-layer perceptrons (MLPs), see Fig.~\ref{fig:principle}B, and Appendix~\aref{sec:architectures} for architectural details.
These networks were optimized during SSL by gradient-descent using the adaptive rate scaling of the LARS algorithm~\citep{you2017large} with learning rate warmup and decay.
If not otherwise stated, SSL was performed for 1,000 epochs.
See Appendix~\aref{sec:optimization} for details to the optimization with SSL and for details of training the linear classifier.
For all our comparisons and ablation studies, we stress that the \textbf{overlap in the implementation is maximal}. Especially for the comparison between SimCLR and S-TEC (and between ReLIC and S-TEC$^*$), we emphasize that only loss functions are changed and manipulation-related inverse models are added.
As mentioned in Section~\ref{sec:recover}, the identity-related inverse model can also be instantiated based on other methods for SSL, such as ReLIC~\citep{mitrovic2021representation} through suitable choice of $q_\theta(a_\mathrm{id}|x,x')$, see Appendix~\aref{sec:derivation}.
We denoted this specific variation with S-TEC$^*$ (i.e.\ target networks etc.).

\paragraph{CIFAR-10/100 and STL-10.}
We report the accuracies that linear classifiers could attain after SSL on the respective datasets in Table~\ref{tab:results}.
Using 5 independent runs for each setting that we considered revealed that the manipulation-related inverse model in the case of S-TEC or S-TEC$^*$ consistently increased the accuracy of a linear classifier over the respective baseline.

\begin{table}[h]
	\caption{Comparing with the results of~\citep{lee2021improving} on STL-10, $\dagger$ see Table 7 thereof.}
	\label{tab:comparing}
	\centering
	\begin{tabular}{lcc}
		\toprule
		& 200 Epochs & 1,000 epochs \\
		\midrule
		SimSiam {\scriptsize impl. by \citet{lee2021improving}} & 86.32$\dagger$ & \textcolor{black}{90.2 {\scriptsize $\pm 0.3$}} \\
		SimSiam + AugSelf {\scriptsize \citep{lee2021improving}} & 86.03$\dagger$ & 90.8 {\scriptsize $\pm 0.2$} \\
		\midrule
		SimCLR {\scriptsize repr. \citep{chen2020simple}} & \textcolor{black}{86.1 {\scriptsize $\pm 0.2$}} & \textbf{91.5} {\scriptsize $\pm 0.4$} \\
		S-TEC {\scriptsize (ours)} & \textcolor{black}{86.2 {\scriptsize $\pm 0.2$}} & \textbf{91.6} {\scriptsize $\pm 0.2$}\\
		\bottomrule
	\end{tabular}
\end{table}

\begin{wraptable}{r}{6.8cm}
	\vspace{-.15cm}
	\caption{ImageNet results (ResNet-50).}
	\label{tab:imagenet}
	\centering
	\begin{tabular}{lc}
		\toprule
		Method (100 epoch)   & Top-1 (val.) \\
		\midrule
		SimCLR {\scriptsize repr. \citep{chen2020simple}} & 64.6 \\
		ReLIC {\scriptsize repr. \citep{mitrovic2021representation}} & 66.2 \\
		S-TEC {\scriptsize (ours)} & 64.8 \\
		S-TEC* {\scriptsize (ours)} & \textbf{66.3} \\
		\midrule
		Method (300 epoch) & Top-1 (val.) \\
		\midrule
		ReLIC {\scriptsize repr. \citep{mitrovic2021representation}} & 70.0 \\
		S-TEC* {\scriptsize (ours)} & \textbf{70.2} \\
		\midrule
		\midrule
		Method ($\geq$ 800 epoch) & Top-1 (test) \\
		\midrule
		MoCo v2 {\scriptsize \citep{chen2020improved}} & 71.1 \\
		SwAV {\scriptsize \citep{caron2020unsupervised}} & 75.3 \\
		SimCLR {\scriptsize \citep{chen2020simple}} & 69.3 \\
		BYOL {\scriptsize \citep{grill2020bootstrap}} & 74.3 \\
		ReLIC {\scriptsize \citep{mitrovic2021representation}} & 74.8 \\
		ReLICv2 {\scriptsize \citep{tomasev2022pushing}} & \textbf{77.1} \\
		VICReg {\scriptsize \citep{bardes2022vicreg}} & 73.2 \\
		\bottomrule
	\end{tabular}
\vspace{.4cm}
\caption{Transfer learning on PASCAL VOC.}
\label{tab:pascal}
\centering
\begin{tabular}{lc}
	\toprule
	\multicolumn{2}{l}{Method (300 epoch) - Obj. Detection (AP50)}\\
	\midrule
	ReLIC {\scriptsize repr.~\citep{mitrovic2021representation}} & 82.3 {\scriptsize (\texttt{test2007})} \\
	S-TEC$^*${\scriptsize (ours)} & \textbf{82.5} {\scriptsize (\texttt{test2007})}\\
	\midrule
	\multicolumn{2}{l}{Method (300 epoch) - Segmentation (mIoU)}\\
	\midrule
	ReLIC {\scriptsize repr. \citep{mitrovic2021representation}} & \textcolor{black}{69.9} {\scriptsize (\texttt{val2012})}\\
	S-TEC$^*${\scriptsize (ours)} & \textcolor{black}{\textbf{70.5}} {\scriptsize (\texttt{val2012})}\\
	\midrule
	\multicolumn{2}{l}{Method (1000 epoch) - Segmentation (mIoU)}\\
	\midrule
	BYOL {\scriptsize \citep{grill2020bootstrap}} & 76.3 {\scriptsize (\texttt{val2012})}\\
	ReLICv2 {\scriptsize \citep{tomasev2022pushing}} & \textbf{77.9} {\scriptsize (\texttt{val2012})}\\
	\bottomrule
\end{tabular}
\vspace{-.6cm}
\end{wraptable}

We also compared our approach with the results of~\citet{lee2021improving}, who considered a similar augmentation-aware training setting, that was mainly focused on the transferability of representations between domains.
We considered the case in which their method exhibited the strongest improvement on STL-10, see Table~6 of~\citep{lee2021improving} (``\texttt{crop}''), and retrained their model using the same number of epochs (1,000).
We used their implementation and employed the same image augmentations as we did.
Results are shown in Table~\ref{tab:comparing}.
Conversely, we also tested our methods in a 200 epoch training budget, as originally done by~\citet{lee2021improving} and included the best reported performance that they obtained, see Table~7 of~\citep{lee2021improving}, noting that our method did not outperform in this case.

To investigate the differences between our approach and that of~\citep{lee2021improving}, we conducted an ablation study exchanging the egocentric action representation that we used with an allocentric one.
In addition, we probed the impact of replacing action classification with L2 regression.
Experiments were performed on CIFAR-100 using ResNet-18s, with Egocentric+Classification yielding \textbf{66.6}\% accuracy over the next best setting Allocentric+L2 Regression with 66.1\%, which was employed by~\citet{lee2021improving} (Fig.~\ref{fig:hp}A and Appendix~\aref{sec:additional}).
We hypothesize that classification affords the model more flexibility in its output distribution, thus it can handle uncertainty of its action prediction better.

\paragraph{ImageNet.}

We experimented with ImageNet ILSVRC-2012~\citep{imagenet} to demonstrate that S-TEC and S-TEC$^*$ also scale.
We performed training for either 100 or 300 epochs and report the results in Table~\ref{tab:imagenet}, confirming that S-TEC is not restricted to small datasets. 

\paragraph{Object detection and semantic segmentation.}
SSL aims to install generally useful representations. Thus, we considered downstream tasks beyond classification: object detection and semantic segmentation on PASCAL VOC~\citep{everingham2010pascal} using Faster R-CNN~\citep{ren2015faster} and fully convolutional networks~\citep{long2015fully}, respectively, along with a ResNet-50 backbone.
We initialized this backbone with the parameters that resulted from SSL on ImageNet for 300 epochs, and then fine-tuned the network on the new task (4 runs with different initializations of remaining parameters).
Results are reported in Table~\ref{tab:pascal}, see also Appendix~\aref{sec:architectures} and~\aref{sec:optimization} for details.

\paragraph{Hyperparameter dependence and learning dynamics.}
To assess the dependence of our results on hyperparameters, we carried out several studies on CIFAR-100 with ResNet-18s: We performed a sweep over $\lambda_\mathrm{manip}$ that scales the impact of the manipulation-related loss, and a sweep over the number of bins $K$ used in the classification for action components.
Performance depends significantly on $\lambda_\mathrm{manip}$, but is less affected by $K$, as long as there are not too many bins (i.e. $K < 10$), see Fig.~\ref{fig:hp}B for results.
Lastly, we also exhibit the loss dynamics and learning progress in Appendix~\aref{sec:optimization_progress}.

\paragraph{Audio (LibriSpeech).}
Finally, we also attempted to improve representation learning in the audio domain.
Specifically, we considered the same data and model as introduced by~\citet{oord2018representation}.
In addition to the time-sensitive identity-related inverse model, as it emerges with CPC, we also added a time-insensitive identity-related inverse model.
This allowed us to achieve \textbf{65.4}\% accuracy on phoneme classification (with frozen features) as opposed to 65.1\% that we obtained with CPC.

\section{Analysis and intuitions}
\label{sec:analysis}

\begin{figure}
	\centering
	\includegraphics[width=13.9cm]{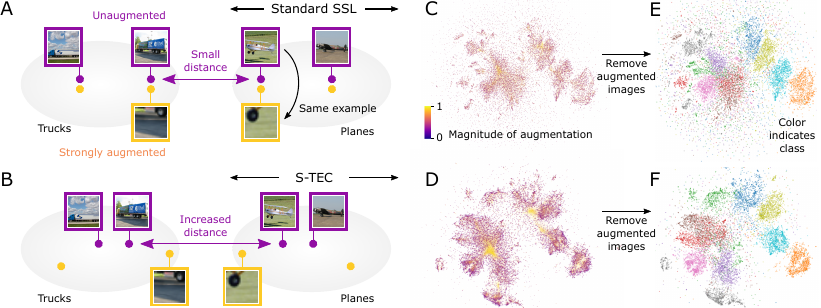}
	\caption{\textbf{Analysis. A-B}) Schematic of conjectured feature space organization: EC-unaware SSL co-locates views of an object (A), while S-TEC separates according to the level of augmentation (B). \textbf{C-D}) UMAP embeddings of images and their augmentations. Collapsed representations become separated. \textbf{E-F}) UMAP embeddings of unaugmented images, using the projection of C-D. Increased class-separation is visible.}
	\label{fig:analysis}
\end{figure}

\paragraph{Increased information content.}
A consequence of learning a manipulation-related inverse model is that additional information must be expressed by the feature extractor $f$.
Curiously, it had been shown in several other works that strong methods for SSL only perform well on downstream tasks, such as classification, if \emph{intermediate} representations are used.
E.g.\ ~\citet{chen2020simple} showed that inserting a nonlinear MLP between the feature extractor $f$ and the loss for contrastive SSL resulted in significantly better performance on subsequent linear classification, as opposed to the control case where this MLP was missing.
This effect is explained by loss of information that is not necessarily important for the contrastive SSL objective, but for downstream tasks.
In fact, recent work~\citep{chen2020big,mitrovic2021representation} observed better performance if the depth of MLP was further increased. This supports the viewpoint that additional information, albeit being potentially redundant to the contrastive SSL objective, is desired for downstream tasks of interest, see also~\citet{lee2021improving}.

\paragraph{Better organization of class borders.}
Furthermore, we conjecture that learning the additional manipulation-related inverse model using $\mathcal{L}_\mathrm{manip}$ for S-TEC (see Eq.~\eqref{eq:general_loss}) encourages the feature space to be better organized.
Firstly, note that conventional (contrastive or non-contrastive) SSL promotes representations of the same object in different views to be co-located~\citep{wang2021understanding} (see Fig.~\ref{fig:analysis}A, e.g.\ truck and its tire).
On the other hand, S-TEC, due to its EC-aware learning, encourages representations of different views of one object (e.g.\ full truck vs tire) to take different positions (truck and tire in Fig.~\ref{fig:analysis}B). As a result, we hypothesize, the representations of canonical, untransformed views of the same type of object (e.g.\ trucks) must become more concentrated, to allow the transformed ones to spread. This must then increase the separation between clusters of untransformed objects (Fig.~\ref{fig:analysis}: purple arrow, A vs B).
Moreover, S-TEC's separation of augmented views from unaugmented ones within an object-class allows the model to instead locate similar augmented views of different object classes. This then forms arguably semantically meaningful class-borders and transitions (Fig.~\ref{fig:analysis}B, truck tire and plane tire).

Experimentally, this hypothesis is supported by the features computed by a ResNet-50 on images of the testing set of CIFAR-10, including also their augmentations.
In Fig.~\ref{fig:analysis}C-F, we computed lower-dimensional projections by the means of UMAP~\citep{mcinnes2018umap}, and colour-coded the magnitude of augmentation (defined as 1 minus the relative area of the cropped image).
Comparing Fig.~\ref{fig:analysis}C and D, the model trained with S-TEC (Panel D) is aware of the zoom level of the augmentation and places more augmented images in similar regions, i.e.\ the borders, while the model trained without the full EC (SimCLR in this case) is oblivious to it (Panel C).
If we embed the original unaugmented images in the same projection (Fig.~\ref{fig:analysis}E-F), the apparent class centre distance increases for S-TEC (Panel F), due to the now missing augmented images on the border.
Quantitatively, we computed for each class separately the distance between the centroid of augmented image representations and the centroid of unaugmented image representations. Averaged over all classes, we find that this distance is 10.5 for S-TEC and 0.4 for SimCLR (both latent spaces cover similar scales). This further confirms that SimCLR clusters these subsets (augmented and unaugmented images) of one class around a single centroid, whereas S-TEC separates them.

Importantly, S-TEC may offer a new method for avoiding \textbf{representational collapse}~\citep{grill2020bootstrap,bardes2022vicreg,balestriero2022contrastive} in non-contrastive SSL, because it explicitly displaces representations of the same object if they correspond to different manipulations. We have shown how S-TEC's theoretical framework recovers non-contrastive learning and extends it with a manipulation-related inverse model, however experiments are left for future work.
\section{Related work}

\paragraph{SSL through auxiliary tasks.}
The idea of SSL by solving high-level queries about input manipulations was considered previously.
E.g.\ \citet{doersch2015unsupervised, noroozi2016unsupervised} proposed to transform input images into patches and attempted to solve context prediction and jigsaw-puzzles respectively, while others found it useful to predict a prior rotation transformation~\citep{gidaris2018unsupervised}.
In contrast to such spatial prediction tasks, a different line of work by~\citet{zhang2016colorful} discovered that colourization of black-and-white images also creates useful features for downstream tasks.
Since aforementioned auxiliary tasks are orthogonal at large, prior works studied combinations and/or extensions of those~\citep{doersch2017multi,zhang2019aet}.

\paragraph{Contrastive SSL.} 
Opposing to the preceding strategies of training on (handcrafted) auxiliary tasks are algorithms that originated from the idea of mutual information (MI) maximization between the input and representations thereof.
In particular, the prevalent strategy of mini-batch training rendered it practical to compare -- and contrast -- representations of different and related inputs, which enables the maximization of MI~\citep{gutmann2010noise}.
Following this perspective, algorithms for deep networks were introduced~\citep{oord2018representation},
and further progress ensued, where the common blueprint for the algorithm is to transform input into pairs and to contrast those against other, unrelated ones~\citep{henaff2020data,he2020momentum,chen2020simple,chen2020big,chen2021intriguing}.
\citet{tschannen2020mutual} opened a discussion whether it works well due to maximization of MI, which resulted in the search for different explanations,~e.g.\ through causal interventions~\citep{mitrovic2021representation,tomasev2022pushing}.
On the other hand, the profound utility of contrastive SSL has also inspired other research that attempts to connect it to hypothesized learning mechanisms in the brain.
For instance,~\citet{illing2021local} show that contrastive SSL can give rise to deep representations with local learning rules.

\paragraph{Non-contrastive SSL.}
Several other studies explored non-contrastive avenues for SSL, which have gained more traction due to their attractive properties, such as not needing negative examples.
In general, these approaches require the network to produce consistent representations under content-preserving input transformations while addressing the problem of representational collapse~\citep{grill2020bootstrap,chen2021exploring,zbontar2021barlow,bardes2022vicreg}.
Furthermore,~\citet{caron2020unsupervised} demonstrate that enforcing consistency between cluster assignments can install potent feature extraction capabilities into a model, while also not requiring pair-wise contrasting.

\paragraph{Augmentation-aware self-supervision.}
Algorithms based on contrastive SSL typically aim for representations that are invariant to input transformations. While this seems appropriate in principle, subsequent studies have shown that this is not always favourable, as this strategy can exclude certain information from representations that could otherwise make them more useful for varying downstream tasks~\citep{xiao2021what}.
Based on similar arguments, \citet{lee2021improving} proposed to predict differences of certain transformation parameters in addition to training on a standard SSL objective~\citep{chen2020improved,chen2020simple,chen2021exploring} to improve the transferability of learned representations to other domains.
These are promising results; however, on the main performance tests of SSL, i.e.\ testing the representations in the same domain as the training domain, these prior augmentation-aware approaches have not achieved the same performance advantage as compared to S-TEC, see Table~\ref{tab:comparing}.

\paragraph{The relation of our approach to prior work.}
Our approach generalizes methods that pair representations of paired inputs into a framework that also introduces semantic structure between paired inputs, based on the known transformations between them.
This generalized and unified framework emerges from the concept of ECs and its relation to inverse models.
Therefore, our approach is augmentation-aware, but its foundation on sensory-motor principles and neuroscience instructs important elements (Section~\ref{sec:stec}) that are missing from earlier augmentation-aware approaches~\citep{xiao2021what,lee2021improving}, but have been discussed analogously by ~\citet{mineault2021your}, and studied in part by research on local learning~\citep{illing2021local}.
Currently, contrastive SSL is one of the dominant approaches in the literature, and our approach improves it (and can be combined with further improvements, e.g.\ ReLIC) in our tests (Tables~\ref{tab:results} and~\ref{tab:imagenet}). Our theoretical framework also recovers and extends non-contrastive approaches, such as BYOL (\Cref{sec:recover}).

\section{Conclusion}
S-TEC is a theoretical framework derived formally from first principles of biological sensory-motor control. It unifies and extends theoretically several SSL approaches, and improves them practically. Interestingly, designing S-TEC's details in a biologically-principled way is crucial for performance. S-TEC is consistently better over several strong baselines in image classification, segmentation, and object detection. S-TEC as a framework provides a new angle for future further improvements to SSL.
By following established biological principles, S-TEC feeds back to neuroscience. Our results suggest that the availability of ECs to the nervous system for inverse-model learning may positively impact sensory skill. This hypothesis is testable. It predicts that subjects, exposed to a motor learning task in a novel sensory environment through active movements, would perceive the new environment better than participants that only experience passive exploration of the environment. Supporting evidence from kittens and humans already exists \citep{held1963movement, bach1972brain}.
S-TEC's biological implications could be strengthened even within the computational setting, by using optimization algorithms that are more biologically plausible than backpropagation. Such options have recently been described, including within SSL~\citep{illing2021local}. Adding further detail to S-TEC's neural networks, such as spiking neurons, could further enhance its biological relevance.
\\\textbf{Limitations.}
Even though our theoretical framework includes and extends various SSL methods, such as the very recent ReLICv2~\citep{tomasev2022pushing}, as well as non-contrastive SSL, e.g.\ BYOL~\citep{grill2020bootstrap}, experimentally we have only extended the methods of~\citet{chen2020simple,mitrovic2021representation, oord2018representation}.
In addition, we have employed only the basic commonly used augmentations for SSL without exploring other actions/augmentations.
Moreover, we have experimented only with ResNets~\citep{he2016deep}.
Future research could test the advantage of S-TEC in other architectures, e.g.\ with self-attention~\citep{vaswani2017attention}.
\\ \textbf{Potential negative societal impacts.}
SSL can exploit unlabelled data, and S-TEC's rich feedback from ECs improves it, thus significantly expanding also the malicious applicability of ML.
One concern is of privacy, i.e.\ S-TEC might assist the profiling of individuals from anonymized data.

\begin{ack}
This work is partially supported by the Science and Technology Innovation 2030-Major Project (Brain Science and Brain-Like Intelligence Technology) under Grant 2022ZD0208700.
The authors would like to thank Lukas Cavigelli, Renzo Andri, Édouard Carré, and the rest of Huawei's Von Neumann Lab, for offering compute resources.
\end{ack}

\bibliographystyle{apalike}
\bibliography{references}


\begin{checklist}
\section*{Checklist}
\begin{enumerate}
	
	\item For all authors...
	\begin{enumerate}
		\item Do the main claims made in the abstract and introduction accurately reflect the paper's contributions and scope?
		\answerYes{}
		\item Did you describe the limitations of your work?
		\answerYes{}
		\item Did you discuss any potential negative societal impacts of your work?
		\answerYes{}
		\item Have you read the ethics review guidelines and ensured that your paper conforms to them?
		\answerYes{}
	\end{enumerate}
	
	\item If you are including theoretical results...
	\begin{enumerate}
		\item Did you state the full set of assumptions of all theoretical results?
		\answerYes{}
		\item Did you include complete proofs of all theoretical results?
		\answerYes{Will be in the supplemental material}
	\end{enumerate}
	
	\item If you ran experiments...
	\begin{enumerate}
		\item Did you include the code, data, and instructions needed to reproduce the main experimental results (either in the supplemental material or as a URL)?
		\answerYes{Will be in the supplemental material}
		\item Did you specify all the training details (e.g., data splits, hyperparameters, how they were chosen)?
		\answerYes{}
		\item Did you report error bars (e.g., with respect to the random seed after running experiments multiple times)?
		\answerYes{}
		\item Did you include the total amount of compute and the type of resources used (e.g., type of GPUs, internal cluster, or cloud provider)?
		\answerNo{}
	\end{enumerate}
	
	\item If you are using existing assets (e.g., code, data, models) or curating/releasing new assets...
	\begin{enumerate}
		\item If your work uses existing assets, did you cite the creators?
		\answerNA{}
		\item Did you mention the license of the assets?
		\answerNA{}
		\item Did you include any new assets either in the supplemental material or as a URL?
		\answerNA{}
		\item Did you discuss whether and how consent was obtained from people whose data you're using/curating?
		\answerNA{}
		\item Did you discuss whether the data you are using/curating contains personally identifiable information or offensive content?
		\answerNA{}
	\end{enumerate}
	
	\item If you used crowdsourcing or conducted research with human subjects...
	\begin{enumerate}
		\item Did you include the full text of instructions given to participants and screenshots, if applicable?
		\answerNA{}
		\item Did you describe any potential participant risks, with links to Institutional Review Board (IRB) approvals, if applicable?
		\answerNA{}
		\item Did you include the estimated hourly wage paid to participants and the total amount spent on participant compensation?
		\answerNA{}
	\end{enumerate}
	
\end{enumerate}
\end{checklist}

\clearpage
\appendix

\renewcommand\thefigure{S\arabic{figure}}    
\renewcommand\thetable{S\arabic{table}}   
\def\theequation{S\arabic{equation}}
\setcounter{figure}{0}
\setcounter{table}{0}
\setcounter{equation}{0}

\newcommand{\mref}[1]{\ref{#1}}

\begin{concatenatedappendix}
	\include{appendix_content}
\end{concatenatedappendix}

\end{document}

%% file: appendix_content.tex
\section{Experimental methods: Input transformations}
\label{sec:input_transformations}
In practice, mini-batch training was performed, hence we applied input transformations to each datapoint twice to ensure that for every $x$, there is always one $x''$ related through $a_\mathrm{id}=0$, i.e.\ having the same underlying object identity.
As a result, using a batch size of $B$ different images will cause $2 B$ images being processed at a time.

We exhibit below the complete list of data augmentation methods in the order that they were applied during SSL.
Note that only \emph{Random crop} and \emph{Random horizontal mirroring} were considered as part of the same-object manipulations $a_\mathrm{manip}$, see Section~\ref{sec:trans-manip}.

\begin{enumerate}
	\item \textbf{Random crop.}
	
	For each transformation we randomly extracted a patch of the image with an area sampled uniformly between 8\% and 100\% of the original area with an aspect ratio sampled log-uniformly between $\frac{3}{4}$ and $\frac{4}{3}$.
	This patch was resized to 32x32, 96x96 or 224x224 pixels for CIFAR-10/100, STL-10 or ImageNet respectively (using bilinear interpolation).
	\item \textbf{Random horizontal mirroring.}
	
	For each transformation we mirrored the image separately with a probability of 50\% horizontally.
	\item \textbf{Random colour jittering.}
	
	With a probability of 80\%, we randomly altered separately for each transformation the brightness, contrast, saturation and hue in a random order.
	More accurately, brightness was adjusted by multiplication of the pixel values with a factor that was uniformly sampled in $[1 - u, 1 + u]$ (I.e.\ multiplicative change in brightness).
	The contrast was adapted by scaling the distance of the pixels from their mean, i.e. $a(x - \mu) + \mu$, where $\mu$ was the average pixel value of the image (weighted according to red: 0.2989, green: 0.587, blue: 0.114), and $a$ was sampled uniformly in $[1 - u, 1 + u]$. 
	The saturation was adapted similarly, but in this case the mean was computed per pixel location. 
	The change in hue was performed in HSV colour space by adding to the H channel a value sampled uniformly in $[-v, v]$ modulo 1.
	
	For CIFAR-10/100 $u$ and $v$ were set to $0.4$ and $0.1$ regardless of the SSL method. For STL-10 and ImageNet experiments $u$ and $v$ were given by $0.8$ and $0.2$ in the cases of SimCLR or S-TEC. In our experiments with ReLIC and S-TEC$^*$, these values were halved.
	\item \textbf{Random conversion to grayscale.}
	
	For each transformation the image was converted separately to grayscale with a probability of 20\%.
	For this conversion the same weighting strategy as described above was employed.
	\item \textbf{Random gaussian blur (Only for STL-10 and ImageNet).}
	
	With probability of 50\%, we applied a Gaussian blur filter separately for each transformation.
	This filter had kernel edge dimensions of 10\% of the image width and height (rounded to uneven edge lengths), and used a standard deviation that was sampled randomly for each transformation uniformly in [0.1, 2.0] for size 224x224 and scaled proportionally in case of other dimensions.
	\item \textbf{Random solarization (Only for STL-10 and ImageNet).}
	
	With a probability of 20\%, we also applied solarization of the image for each transformation separately.
	This was performed by inverting pixels with a value above 0.5 (assuming a pixel value range of 0 to 1).
	Here, inversion refers to a mapping $x \mapsto 1 - x$.
	
	Note that we \textbf{excluded} the loss for the manipulation-related inverse model $\mathcal{L}_\mathrm{manip}$ if either $x$ or $x'$ had been solarized.
\end{enumerate}

\subsection{Object-identity-related actions $a_\mathrm{id}$}
Since we applied training in mini-batches with $B$ different images, the action $a_\mathrm{id}=1$ is simply given by $x$ and $x'$ corresponding to different image identities in the batch.

\subsection{Same-object manipulations $a_\mathrm{manip}$}
\label{sec:trans-manip}
The same-object manipulations $a_\mathrm{manip}$, as introduced in Section~\mref{sec:definitions} in the main manuscript, are only applied if the object identity of $x$ and $x'$ remains the same, which is the case when $a_\mathrm{id}=0$.

The same-object manipulations $a_\mathrm{manip}$, as defined in our setting, took into account only spatial operations: \emph{Random crop} and \emph{Random horizontal mirroring}.
To represent this action, we first computed for each transformation the affine matrix $M_x$ that generates the particular cropped view of $x$ from the original image.
More specifically, this matrix $M_x$ transforms points on the canvas of the new view $x$ to the points on the canvas of the original image, i.e.\ $M_x$ determines the source position of the new pixels.

To compute $M_x$, let $w_x$ and $h_x$ denote the width and height of the crop in pixels as sampled from the \emph{Random crop} operation.
In addition, let $l_x$ ($t_x$) be the distance of the crop's left (top) edge from the original image's left (top) edge. 
Additionally, let $W$ and $H$ denote the width and height of the original image respectively.
Furthermore, let $f_x$ be -1 if the \emph{Random horizontal mirroring} operation dictates a mirroring and 1 if not.
With these definitions $M_x$ is defined by:
\begin{align}
	M_x = \begin{pmatrix}
		f_x \frac{w_x}{W} & 0 & \frac{w_x}{W} - 1 + 2 \frac{l_x}{W} \\
		0 & \frac{h_x}{H} & 1 - \frac{h_x}{H} + 2 \frac{t_x}{H} \\
		0 & 0 & 1
	\end{pmatrix} =: \begin{pmatrix}
		m_x^{(1,1)} & m_x^{(1,2)} & m_x^{(1,3)} \\
		m_x^{(2,1)} & m_x^{(2,2)} & m_x^{(2,3)} \\
		m_x^{(3,1)} & m_x^{(3,2)} & m_x^{(3,3)}
	\end{pmatrix}~.
\end{align}
Since we identify with $a_\mathrm{manip}$ the action that turns $x$ into $x'$, we are interested in the affine transformation matrix $M_{x\rightarrow x'}$ that transforms $x$ to $x'$ (i.e.\ it computes the source location of pixels in $x'$ on the canvas of $x$).
It is given by:
\begin{align}
	M_{x\rightarrow x'} = M_{x'} M_x^{-1}~.
\end{align}
Finally, we identify the spatial action $a_\mathrm{manip}$ with the two top rows of this matrix:
\begin{align}
	a_\mathrm{manip} = (m_{x\rightarrow x'}^{(1,1)}, m_{x\rightarrow x'}^{(1,2)},m_{x\rightarrow x'}^{(1,3)},m_{x\rightarrow x'}^{(2,1)},m_{x\rightarrow x'}^{(2,2)},m_{x\rightarrow x'}^{(2,3)})
\end{align}

\paragraph{Categorical targets.} In order to classify the values of the matrix of $a_\mathrm{manip}$, we subdivided the interval of values that can be assumed into $K=6$ bins.
For that we first define limits ${\mathrm{manip}_\mathrm{min}=(-2,-2,-0.5,-2, -2,-0.5)}$ and ${\mathrm{manip}_\mathrm{max}=(2,2,0.5,2, 2,0.5)}$, which ultimately allows us to express the discretized $\widehat{a}_{k, \mathrm{manip}}$ of the main manuscript in Section~\mref{sec:formalism} as:
\begin{align}
	\widehat a_{k, \mathrm{manip}} = \max \left(  \min \left( \Big\lfloor \frac{a_{k,\mathrm{manip}} - \mathrm{manip}_{k,\mathrm{min}}}{\mathrm{manip}_{k,\mathrm{max}} - \mathrm{manip}_{k,\mathrm{min}}} \Big\rfloor  ,K-1\right) , 0\right)~.
\end{align}

\section{Experimental methods: Architectures}
\label{sec:architectures}
The architectures for feature encoder $f$ were residual convolutional networks as introduced by~\citet{he2016deep} (i.e. ResNet v1). 
More specifically, we used ResNet-18 or ResNet-50, depending on the experiment, and used the activations after global average pooling as the output of $f$.

The functions $\phi$, for the identity-related inverse model, and $\psi$, for the manipulation-related inverse model, were based on multilayer perceptrons (MLPs) with batch normalization and rectified linear activation (ReLU).

\subsection{Identity-related inverse model $\phi$.}
\label{sec:arch-identity-related-inverse-model}
The function $\phi$ was defined by a cosine similarity of the outputs of an MLP $g$:
\begin{align}
	\phi(a, b) = \sum_j \frac{g_j(a)}{\|g(a)\|_2} \frac{g_j(b)}{\|g(b)\|_2}~,
\end{align}
where the MLP $g$ had 1 hidden layer with batch normalization and ReLU activations.
Batch normalization was also used for its output (except when target networks were used in S-TEC*).
The number of hidden and output units of $g$ differed among experiments, see Table~\ref{tab:g_sizes} for concrete dimensions.
\begin{table}
	\caption{Parameters of the MLP $g$ per learning experiment.}
	\label{tab:g_sizes}
	\centering
	\begin{tabular}{llccc}
		\toprule
		Architecture for $f$ & Parameters of $g$     & CIFAR-10/100     & STL-10 & ImageNet \\
		\midrule
		\multirow{2}{*}{ResNet-18} &
		Hidden size & 512 & 512 & - \\
		& Output size & 64 & 128 & - \\
		\midrule
		\multirow{2}{*}{ResNet-50} &
		Hidden size & 2,048 & 2,048 & 2,048 \\
		& Output size & 64 & 128 & 128 \\
		\bottomrule
	\end{tabular}
\end{table}

\subsection{Manipulation-related inverse model $\psi$.}
The manipulation-related inverse model $\psi$ was defined in the main manuscript as a function of two feature vectors.
It was implemented as an MLP, applied on the concatenation of both inputs, with one hidden layer that contained 512 units with batch normalization and ReLU activation.
The output of $\psi$ was 36 dimensional in total, producing predictions for each component $a_{k,\mathrm{manip}}$, in which of the $K=6$ bins its value falls.

\subsection{S-TEC*}
We also experimented with ReLIC~\citep{mitrovic2021representation}, which modifies the approach by introducing target networks and an overall confidence factor $\exp(-\alpha D_{c1,c2})$, which is explained in Section~\ref{sec:relic_recovery}.
In this case, the definition of $q_\theta$ becomes:
\begin{align}
	q_\theta(a_\mathrm{id}=0|x,x',\theta) = \frac{\exp( \tilde\phi (f(x), \tilde f(x') ) / \tau )}{\sum_{x_n \in \{x'\}\cup C} \exp( \tilde\phi (f(x), \tilde f(x_n) ) / \tau )} \exp(-\alpha D_{c1,c2})~,
\end{align}
where we used $\tilde f$ as a target network that follows the weights of $f$ using an exponential moving average~\citep{mitrovic2021representation} with same decay properties as in~\citep{grill2020bootstrap} using an inital decay $\tau=0.99$.
In addition, the exponential moving average was also applied to the MLP $g$ such that ${\tilde \phi(a, b) = \phi(a, b) = \sum_j \frac{g_j(a)}{\|g(a)\|_2} \frac{\tilde g_j(b)}{\|\tilde g(b)\|_2}}$, with $\tilde g$ following the weights of $g$ using an exponential moving average.

\subsection{MoCo}
For our implementation of MoCo v2~\citep{chen2020improved}, we used target networks $\tilde f$ and $\tilde g$ with the same exponential moving average schedule as used by~\citep{grill2020bootstrap} with an initial decay of $\tau=0.99$. All other architectural settings were kept equal to the SimCLR setting, see Table~\ref{tab:g_sizes}. The size of the dictionary, i.e. the bank of contrastive embeddings, was set to 65k.

\subsection{BYOL}
For our implementation of BYOL, we followed the architectural principles as provided by~\citet{grill2020bootstrap}. However, for CIFAR-10/100 we used a hidden dimension of 512 for projection and predictor, as well as an output dimensionality of 64. For STL-10, we used a hidden dimension of 2048 with an output dimensionality of 128.

\subsection{Object detection}
For object detection, we employed Faster R-CNN~\citep{ren2015faster} with a ResNet-50 backbone.
In general, we followed the architectural settings of~\citep{he2020momentum}, but adopted an additional batch normalization layer not only for the box prediction head, but also for the region proposal network (RPN) just before the linear output layers.

\subsection{Semantic segmentation}
For semantic segmentation, we employed fully convolutional networks~\citep{long2015fully} with a ResNet-50 backbone.
More specifically, we followed the settings of~\citep{he2020momentum}, where we retain only convolutional layers of the ResNet, replacing stride in the last convolution block (conv$_5$) with a dilation of 2.
After that, two 3x3 convolutions, each with batch normalization and ReLU activation, are added, followed by a 1x1 convolution for pixel-wise classification.
This design yields a total stride of 16 (FCN-16s~\citep{long2015fully}).

\section{Experimental methods: Optimization}
\label{sec:optimization}

\subsection{SSL phase}
We used stochastic gradient-descent with a momentum of 0.9 along with the LARS adaptive learning rate mechanism~\citep{you2017large}, but excluded batch normalization and bias parameters from it.
We used a batch size $B$ of 1024 for all our experiments, except for those on ImageNet, where we used a $B = 1680$.
Recall that $B$ denotes the number of different images, each of which was subject to 2 augmentations, resulting in $2B$ images processed at a time.

We employed linear scaling of the learning rate with respect to the batch size, with cosine learning rate decay and with 10 epochs of linear warmup, see Table~\ref{tab:hp_opt} for learning rates per 256 batch size.
Global weight decay was used as part of $\mathcal{L}_\mathrm{reg}$ with a coefficient of $10^{-6}$.

Specifically for the ResNet, we note that the last batch normalization layer in each residual block was initialized with zero scale to stabilize training~\citep{goyal2017accurate}.

\begin{table}[h]
	\caption{Optimization hyperparameters per learning experiment.}
	\label{tab:hp_opt}
	\centering
	\begin{tabular}{lccc}
		\toprule
		Hyperparameter & CIFAR-10/100     & STL-100 & ImageNet \\
		\midrule
		Learning rate per 256 batch size & 1.0 & 0.3 & 0.3 \\
		Temperature $\tau$ & 0.5 & 0.2 & 0.1 \\
		Coefficient $\lambda_\mathrm{manip}$ (S-TEC) & 1.0 & 0.3 & 0.6 \\
		\bottomrule
	\end{tabular}
\end{table}

\subsection{Linear classification on frozen features}
For classification we trained a linear classifier on top of the frozen features using stochastic gradient descent.
We trained this linear classifier alongside SSL training in the cases of CIFAR-10/100 and ImageNet, but without propagating gradients into the ResNet feature extractor $f$ (i.e.\ we used \texttt{stop\_gradient} for the classification loss), noting that similar results were achieved with a subsequent optimization protocol consistent with~\citep{chen2020simple}.

Specifically for STL-10, we trained the linear classifier separately in a subsequent optimization procedure with stochastic gradient descent and Nesterov momentum of 0.9, using a learning rate of 0.01 per 256 batch size (we found the value of 0.175 to work best in the case of BYOL and MoCo).
In this case, only random cropping and random horizontal mirroring were used as augmentation methods.
This optimization program was carried out for 2,000 epochs using cosine learning rate decay and 10 epochs of warmup.
For the weights of the linear classifier (excluding bias), a weight decay regularization with a coefficient of $5 \cdot 10^{-4}$ was used.

\subsection{Object detection}
For object detection on PASCAL VOC, we largely followed the settings of~\citep{he2020momentum} and fine-tuned network parameters end-to-end, with training data from the \texttt{trainval2007+2012} splits, while evaluation was carried out on \texttt{test2007}.
Training was performed for 24K iterations with stochastic gradient descent (using a momentum of 0.9) and a batch size of 15.
The learning rate was set to 0.7, which was linearly warmed up for 1K iterations, and then multiplied by 0.1 at 18K and 22K iterations.
The loss coefficient for region proposal network-related losses was set to 0.2.
No weight decay was employed.

\subsection{Semantic segmentation}
For semantic segmentation on PASCAL VOC, we also largely followed the settings of~\citep{he2020momentum}, where training was performed on an augmented split \texttt{train\_aug2012}, introduced by~\citep{hariharan2011semantic} for 45 epochs with stochastic gradient descent (using a momentum of 0.9) and a batch size of 16.
The learning rate was set to 0.03 (0.003 for ResNet parameters initialized from SSL), which was multiplied by 0.1 at the 70\% progress mark and the 90\% progress mark.
A weight decay of $10^{-4}$ was employed.

\section{Derivations and additional theory}
\label{sec:derivation}

\subsection{Sketch of the derivation}
We begin from the definition that poses the optimization of the inverse model as a minimization of the Kullback-Leibler divergence as defined in Eq.~(\mref{eq:principle}) of the main manuscript.
Based on this and on biologically-inspired assumptions (see Section~\mref{sec:ec} of the main manuscript), the inverse model becomes a classifier of pairs of sensory inputs into actions.

Since we introduced the action as being composed of two categories, namely \emph{Object-identity-related actions} and \emph{Same-object manipulations} (see Fig.~\ref{fig:actions}),
we can decompose the loss into a sum of two losses, $\mathcal{L}_\mathrm{id}$ and $\mathcal{L}_\mathrm{manip}$ respectively (see Section~\ref{sec:decomposition}).
This, in turn, allows us to learn the two associated inverse models separately.

Subsequently, we elaborate on the assumption that this model represents an embodied natural setting.
From there, we show that different SSL methods emerge from S-TEC, depending on the specifics of the classifier's mathematical definition and the corresponding EC-based learning. 
The methods we recover include existing and proven ones, such as SimCLR~\citep{chen2020simple}, BYOL~\citep{grill2020bootstrap}, or ReLIC~\citep{mitrovic2021representation}.

\subsection{Concretizing the inverse model as a classifier}
\label{sec:concretizing_class}
The Kullback Leibler divergence loss of the inverse model (Eq.~(\mref{eq:principle}) of the main manuscript) is computed between (a) a probability distribution over true actions, which are copied through EC (given pairs of sensory inputs), and (b) the modelled probability distribution over actions, which is provided by the inverse model (given pairs of sensory inputs).
Grounded on biological evidence (see Section~\mref{sec:ec} of the main manuscript), we assume that the probability distribution of the EC reflects a perfect copy, and that the probability distributions involved are discrete.

The inverse model estimates which actions were executed that caused the sensory inputs to change from $x$ to $x'$.
Based on biological evidence (see Section~\mref{sec:ec} of the main manuscript), we chose to represent this inverse model as a classifier $q_\theta(a|x,x')$ (parametrized by $\theta$) that assigns probability to specific actions, given the inputs $x$ before the action was executed, and the inputs $x'$ after the action's execution.

We identified two categories of actions in Section~\mref{sec:definitions} of the main manuscript, and therefore chose to view actions as being composed of two components $a=(a_\mathrm{id}, a_\mathrm{manip})$.
As a result, two sub-inverse models (sub-classifiers) can be defined:
\begin{itemize}
	\item the identity-related inverse model $q_\theta(a_\mathrm{id}|x,x')$, and
	\item the manipulation-related inverse model $q_\theta(a_\mathrm{manip}|a_\mathrm{id}, x,x')$
\end{itemize}
by means of ${q_\theta(a|x,x')=q_\theta(a_\mathrm{id}|x,x') q_\theta(a_\mathrm{manip}|a_\mathrm{id}, x,x')}$, see also Fig.~\mref{fig:actions}.
In turn, in order to install the function into this classifier, we defined a loss: the KL divergence in Eq.~(\mref{eq:principle}), which can be decomposed into a sum of two divergences, according to the aforementioned factorization.
See Section~\ref{sec:decomposition} for proof that the decomposition is equivalent to the original loss.
This allows us not only to specify the inverse model as two separate classifiers, but also to learn them separately.

\paragraph{Identity-related inverse model}
The identity-related inverse model has to effectively solve a binary classification problem for $a_\mathrm{id}$ to identify whether the main object of the sensory inputs $x$ is the same as the main object of the subsequent sensory input $x'$ (in which case $a_\mathrm{id}=0$).
The probability that is assigned to this event is denoted by:
\begin{align}
	q_\theta(a_\mathrm{id}=0|x,x', \theta)~.\label{eq:appendix_sw}
\end{align}
The specific implementation of the identity-related inverse model (i.e.\ the classifier for $a_\mathrm{id}$) can use a variety of criteria to determine whether two sensory inputs represent the same object identity.
It is typically based on the similarity of the sensory inputs (more accurately the representations thereof).
Examples of such (dis-)similarity measures include \textbf{(a) dot product}, \textbf{(b) mean squared error}, \textbf{(c) KL-divergence}, and more could be envisioned.

\paragraph{Manipulation-related inverse model}
As commented in the main manuscript, the manipulation-related inverse model has to infer which manipulations were performed, when the same underlying object remains in focus, i.e.\ $a_\mathrm{id}=0$.
Also this inverse model can be conceived in various forms, depending on which interactions and manipulations are possible for the same object.
For manipulation of static images, an example is to classify the components of affine transformation matrices as we proposed in Section~\ref{sec:trans-manip}.

\subsection{Concretizing the embodied natural setting}
\label{sec:concretizing_nat}
To account for a generic and natural setting, we assume that the inverse model's parent entity, i.e.\ the observer or agent that performs the actions, may have contextual information from the environment in addition to the observed $x$ and $x'$.
More specifically, we assume that in addition to the perception of $x$ as the main object, the agent also perceives some additional context $C_\mathrm{pre}$.
Likewise, after execution of the action $a$, we assume that the agent perceives $x'$ and some additional context $C$.
In summary, the agent perceives the set $\{x\} \cup C_\mathrm{pre}$ before executing the action $a$, and $\{x'\} \cup C$ thereafter.
We further specify that in the case where $a_\mathrm{id}=0$ the identity of $x$ and $x'$ is the same.
On the other hand, in the case that the agent switches the focus to a different main object ($a_\mathrm{id}=1$), then the identity of the object $x$ remains in the broader context of the agent and is still represented in the set $C$ in some form.

We additionally assume that the agent has prior knowledge about the physical environment's conservation laws, i.e.\ that one object cannot take more than one identity, and that objects do not vanish without cause.
Altogether, this prior knowledge imposes a constraint on the probabilities of $x$ having the same identity as some other $x_n$ in the context.
More accurately, we say the identity of $x$ must be conserved in the set $\{x'\} \cup C$ after taking the action:
\begin{align}
	\sum_{x_n \in \{x'\} \cup C} q_\theta(a_\mathrm{id}=0|x,x_n) = 1~, \label{eq:conservation}
\end{align}
which is imposed on the identity-related inverse model.

As we will show, using the entire context $\{x'\} \cup C$ during learning or only $x'$, determines if the emerging SSL approach with S-TEC belongs to the contrastive category of methods or not.

\subsection{Learning the identity-related inverse model for $a_\mathrm{id}$}
\subsubsection{Using the context of an object: Contrastive SSL}
\label{sec:using-context}
When the entire context $\{x'\} \cup C$ is available during learning, utilizing the assumptions and emerging constraints from Section~\ref{sec:concretizing_nat}, we arrive at two implications:
\begin{enumerate}
	\item Learning the identity-related inverse model $q_\theta(a_\mathrm{id}|x,x')$ consists in maximizing the probability in Eq.~\eqref{eq:appendix_sw}, if the EC dictates that $x$ and $x'$ share the same object identity.
	On the other hand, if the EC dictates that the identity of $x$ does not match the identity of $x'$, the probability in Eq.~\eqref{eq:appendix_sw} is to be minimized.
	
	\item Consider specifically the case where $x$ and $x'$ do not share the same identity.
	It was assumed that the original $x$ stays preserved in the context $C$ in some, possibly altered, form.
	We denote this preserved item sharing the same identity by $x'' \in C$.
	Through conservation in Eq.~\eqref{eq:conservation} it follows that maximizing the probability ${q_\theta(a_\mathrm{id}=0|x,x'')}$ has as a result the minimization of the probability $q_\theta(a_\mathrm{id}=0|x,x_n)$ for all other ${x_n \in \{x'\} \cup C \setminus \{x''\}}$ ``negative'' objects in the context, therefore explicit separate minimization for the negative examples is not necessary.
\end{enumerate}
Furthermore, this type of learning can use directly the conservation in Eq.~\eqref{eq:conservation} for mutual comparison of $x$ with the items inside the set $\{x'\} \cup C$, typically through the use of normalization of similarities (e.g.\ ``softmax'' of similarity scores), such as how it was defined in Eq.~(\mref{eq:sw}) of the main manuscript.

We refer to Section~\ref{sec:upper_bound} for a concrete proof of the second implication, following the same idea that explicit minimization for ``negative'' objects is not necessary, which then connects this to the upper bound objective in Eq.~(\mref{eq:objective}) in the main manuscript.

\subsubsection{Not using the context of an object: Non-contrastive SSL}
\label{sec:not-using-context}
If contextual objects (see Section~\ref{sec:concretizing_nat}) are unavailable, then learning the identity-related inverse model can still be implemented by a formulation of the probability in Eq.~\eqref{eq:appendix_sw}, solely on the basis of the similarity between the representations of $x$ and $x'$, without a comparison to the context.

That choice therefore implies a non-contrastive type of SSL. By further specifying the options of the realization of this model, we show concretely in Section~\ref{sec:byol_recovery} that existing non-contrastive SSL methods emerge~\citep{grill2020bootstrap}.

In this non-contrastive setting, where the task is to maximize the similarity between paired sensory representations, a trivial solution could be found, where \textit{all} objects collapse to the same representation ~\citep{grill2020bootstrap}.
As a result, a potentially trivial solution can occur, where \textit{all} objects collapse to the same representation, thus maximizing the similarity of all possible representations. 

This has been recognized and it has been shown that such trivial solution can be mitigated by using separate feature extractors for the two representations, and by optimizing them differently by learning in different timescales (i.e.\ ``online networks'' and ``target networks'')~\citep{grill2020bootstrap}.
We conjecture that this complexity and its drawbacks are potentially not necessary when a complete EC is employed through S-TEC, since an additional classification task involving $a_\mathrm{manip}$ must be solved that would naturally prevent such collapse, since it demands separation between representations of differently manipulated views.

\subsection{Learning the manipulation-related inverse model for $a_\mathrm{manip}$}
On the other hand, we have established that in addition to the identity-related inverse model, there also exists the manipulation-related inverse model that classifies which manipulations $a_\mathrm{manip}$ were applied to an object, if the identity of $x$ and $x'$ remains the same.

The learning procedure of the manipulation-related inverse model depends on the specific definition of the model, as well as the type of manipulation actions $a_\mathrm{manip}$ that it models.
In this work, we considered $a_\mathrm{manip}$ as being composed of the components of affine transformation matrices (see main manuscript's Section~\mref{sec:ec} for the motivation, and Sections~\ref{sec:trans-manip} and~\ref{sec:concretizing_class} for details).
The corresponding inverse model for $a_\mathrm{manip}$ was implemented as a classifier that predicts in which bin, i.e.\ class, the components of the actions fell, see Section~\ref{sec:trans-manip}.

As a result, learning the manipulation-related inverse model consisted of training multiple classifiers for all the components of $a_\mathrm{manip}$, in other words minimization of the cross-entropy loss between the target classes in which the components of the manipulation action $a_\mathrm{manip}$ fell and the predicted classes for these components.

\subsection{Simultaneous training of two inverse models for $a_\mathrm{id}$ and $a_\mathrm{manip}$}
Ultimately we aim to train both of these inverse models simultaneously, which could be naively carried out by simply adding together the corresponding losses.
However, in practice, the tasks of the two inverse models can differ in their difficulty, and thus require a different weighting to enable learning of both simultaneously.
For this reason we have introduced a weighting factor $\lambda_\mathrm{manip}$ that scales the impact of the loss concerning the manipulation-related inverse in relation to the one corresponding to the identity-related inverse model, see Fig.~\ref{fig:hp}B for a sweep over this parameter.

Furthermore, since our main goal is to achieve the best possible sensory representations installed in one model, we want to share parts of the architecture for both inverse models regarding $a_\mathrm{manip}$ and $a_\mathrm{id}$.
A direct consequence from doing so is that there may be an interaction between parts of the representation space relating to $a_\mathrm{manip}$ and other parts of the representation space relating to $a_\mathrm{id}$, which we briefly elaborated on in Section~\mref{sec:analysis} of the main manuscript.

Another conjecture, as pointed out in Section~\ref{sec:not-using-context}, is that the presence of the loss relating to $a_\mathrm{manip}$ in addition to the loss relating to $a_\mathrm{id}$ could help to prevent collapse of representations, although we have not tested this hypothesis.
In a similar vein, the particular point at which the two inverse models extract the respective features for their further use, and their depth, may have significant impact on the organization of representations.

\subsection{Recovering prior SSL techniques from S-TEC}
\subsubsection{Recovering SimCLR}
\label{sec:ntxent_recovery}
In case that the entire context $\{x'\} \cup C$ of objects is available and used during learning, we can obtain the NT-Xent type of loss as used in SimCLR~\citep{chen2020simple}.
This emerges from a specific choice for modelling the identity-related inverse model's inferred probabilities
$q_\theta(a_\mathrm{id}=0|x,x_n)$, as follows.

By using the dot-product similarity between the representations of $x$ and each example $x_n \in \{x'\} \cup C$ in the context, and then applying a ``softmax'' operation to convert these values into a probability, yields Eq.~(\mref{eq:sw}) of the main manuscript as the inverse model for $a_\mathrm{id}$.
In addition, statement (2) of Section~\ref{sec:using-context} is employed, which poses the learning of $q_\theta(a_\mathrm{id}|x,x')$ as a maximization of $q_\theta(a_\mathrm{id}=0|x,x'')$ for a $x''\in \{x'\} \cup C$ that shares the same identity as $x$.
This maximization of the softmax probability for the ``positive'' pair is equivalent to minimizing its negative logarithm, which is, in fact, the NT-Xent loss.
In Section~\ref{sec:upper_bound} we also show formally that NT-Xent optimizes an upper bound to the original Kullback Leibler divergence objective for learning the identity-related inverse model.
Thus, we have recovered NT-Xent~\citep{chen2020simple} as a special case of our S-TEC framework, which has been the core mechanism in some of the best performing methods for SSL~\citep{chen2020big,chen2021intriguing}.

\subsubsection{Recovering non-contrastive SSL (BYOL)}
\label{sec:byol_recovery}
In the case where no context is available or used during learning of the identity-related inverse model, the ``BYOL'' approach~\citep{grill2020bootstrap} can be recovered if the identity-related inverse sub-model of S-TEC is realized differently.

Specifically, in order to arrive at the approach of~\citet{grill2020bootstrap}, we begin by defining our identity-related inverse model $q_\theta(a_\mathrm{id}|x,x')$ in accordance to a normal distribution, such that its optimization will result in a mean squared error loss, which is what is used in BYOL.
Namely, we define:
\begin{align}
	q_\theta(a_\mathrm{id}=0|x,x') = (1 - \epsilon) \exp \left(- \| g(f(x)) - \tilde g(\tilde f(x')) \|_2^2 \right)~,
\end{align}
where we have introduced $\tilde f$ and $\tilde g$ to indicate that these networks can be different from $f$ and $g$ but related. In~\citet{grill2020bootstrap}, they are related to the original network $f$ and $g$ via an exponential moving average (target networks).
The constant $\epsilon$ denotes a small number.

We then define that the probability assigned to $a_\mathrm{id}=1$ is a small constant.
Since the probability will generally not sum to 1 for these two cases, we formally introduce $a_\mathrm{id}=2$ that does not occur in practice, i.e.\ ${p_\mathrm{EC}(a_\mathrm{id}=2|x,x') = 0}$:
\begin{align}
	q_\theta(a_\mathrm{id}=1|x,x') &= \epsilon~,\\
	q_\theta(a_\mathrm{id}=2|x,x') &= 1 - q_\theta(a_\mathrm{id}=0|x,x') - q_\theta(a_\mathrm{id}=1|x,x')~.
\end{align} 
Inserting these definitions into the loss of S-TEC's identity-related inverse model, $\mathcal{L}_\mathrm{id}=D_\mathrm{KL}(p_\mathrm{EC}(a_\mathrm{id}|x,x'); q_\theta(a_\mathrm{id}|x,x'))$ (see Section~\ref{sec:decomposition}), recovers the approach of~\citep{grill2020bootstrap}:
\begin{align}
	\mathcal{L}_\mathrm{id} =& D_\mathrm{KL}(p_\mathrm{EC}(a_\mathrm{id}|x, x'); q_\theta(a_\mathrm{id}|x,x')) \nonumber\\
	=& \,\mathrm{const} - \sum_{s=0}^2 p_\mathrm{EC}(a_\mathrm{id}=s|x,x') \log q_\theta(a_\mathrm{id}=s|x,x') \nonumber\\
	=& \,\mathrm{const} + p_\mathrm{EC}(a_\mathrm{id}|x,x') \|g(f(x)) - \tilde g(\tilde f(x')) \|_2^2 (1 - \epsilon)~.
\end{align}

Therefore, BYOL has emerged as another special case of S-TEC.

\subsubsection{Recovering ReLIC and ReLICv2}
\label{sec:relic_recovery}
Finally, we hypothesize that the approach for ReLIC~\citep{mitrovic2021representation} along with its assorted invariance penalty can also be recovered from our framework if one postulates that \textbf{both} contexts $\{x\} \cup C_\mathrm{pre}$ and $\{x'\} \cup C$ (see Fig.~\ref{sec:concretizing_nat}) are available and used during learning. The same principles form the basis of the more recent ReLICv2~\citep{tomasev2022pushing}.

In this case, the idea to arrive there is to base the classifier $q_\theta(a_\mathrm{id}=0|x,x')$ on two factors:
\begin{enumerate}
	\item The probability of the context-aware model that was used to obtain NT-Xent, see Section~\ref{sec:using-context}, and~\ref{sec:ntxent_recovery},
	\item An overall confidence of the identity-related inverse model that is defined based on the consistency between the contexts ${\{x\} \cup C_\mathrm{pre}}$ and ${\{x'\} \cup C}$.
\end{enumerate}
Consider specifically the second point that is added on top of what we considered already in the case of SimCLR in~\ref{sec:ntxent_recovery}.
For the purposes of this derivation, we refer to the identity-related inverse model that emerges for obtaining SimCLR, see the definitions in~\ref{sec:ntxent_recovery} and in~\ref{sec:upper_bound}, as $q^{(\mathrm{NT})}_\theta(a_\mathrm{id}|x,x')$.

Furthermore, we define probability distributions $q_{c1,\theta}(x_1) = q^{(\mathrm{NT})}_\theta(a_\mathrm{id}|x,x_1)$ with $x_1\in \{x'\}\cup C$ and $q_{c2,\theta}(x_2) = q^{(\mathrm{NT})}_\theta(a_\mathrm{id}|x'',x_2)$ with $x_2 \in \{x\} \cup C_\mathrm{pre}$ in order to cross-compare probability assignments between the \textbf{same} objects in both contexts (recall that $x''\in C$ denotes the item with the same identity as $x'$).
The consistency between these distributions is quantified by a Kullback-Leibler divergence $D_\mathrm{KL}(q_{c1,\theta}(x_1);q_{c1,\theta}(x_2))=:D_{c1,c2}$, and is used as an overall confidence for the predictions of the inverse model in the following way:
\begin{align}
	q_\theta(a_\mathrm{id} &= 0|x,x',C_\mathrm{pre},C) = q^{(\mathrm{NT})}_\theta(a_\mathrm{id}=0|x, x') \exp \left( - \alpha D_{c1,c2}\right)~,\nonumber\\
	q_\theta(a_\mathrm{id} &= 1|x,x',C_\mathrm{pre},C) = q^{(\mathrm{NT})}_\theta(a_\mathrm{id}=1|x, x') \exp \left( - \alpha D_{c1,c2}\right)~,\label{eq:relic_inverse_model}
\end{align}
while, similar to~\ref{sec:byol_recovery}, we add a $a_\mathrm{id}=2$ that does not occur in practice such as to absorb the remaining probability: $q_\theta(a_\mathrm{id}=2|\dots) = 1 - q_\theta(a_\mathrm{id}=0|\dots) - q_\theta(a_\mathrm{id}=1|\dots)$. Note that $\alpha$ is a hyperparameter.

It remains to substitute these definitions into $\mathcal{L}_\mathrm{id}$, which yields:
\begin{align}
	\mathcal{L}_\mathrm{id} =& D_\mathrm{KL}(p_\mathrm{EC}(a_\mathrm{id}|x, x'); q_\theta(a_\mathrm{id}|x,x')) \nonumber\\
	=& \,\mathrm{const} - \sum_{s=0}^2 p_\mathrm{EC}(a_\mathrm{id}=s|x,x') \log \left( q^\mathrm{(NT)}_\theta(a_\mathrm{id}=s|x,x') \exp(-\alpha D_{c1,c2}) \right) \nonumber\\
	=& \,\mathrm{const} + \alpha D_{c1,c2} - D_\mathrm{KL}(p_\mathrm{EC}(a_\mathrm{id}|x, x'); q^\mathrm{(NT)}_\theta(a_\mathrm{id}|x,x'))~.
\end{align}
Thus, we obtain the ReLIC method including its consistency loss~\citep{mitrovic2021representation} by suitable definition of the inverse model in Eq.~\eqref{eq:relic_inverse_model}.

\subsection{Summary: S-TEC as a generalization of SSL methods}
From first principles of sensory-motor control in Neuroscience, and the assumption that learning occurs in the physical world, we recovered prior SSL methods. However the full S-TEC model is broader, as it also includes $a_\mathrm{manip}$ in its inverse model, which is not exploited by the methods we recovered through $a_\mathrm{id}$. In the main manuscript's Section~\mref{sec:stec}, we showed that $a_\mathrm{manip}$ is part of the same framework, and, in our experiments and analyses in the other sections of the main manuscript, we showed that it is actually useful to combine the two, if implemented according to ECs and sensory-motor principles.

Moreover, from S-TEC's framework, other powerful instantiations can be imagined. For example, we have mentioned that possibly non-contrastive approaches without a target network could become functional, by avoiding representation collapse, through $a_\mathrm{manip}$. Further SSL concepts emerge by implementing S-TEC's elements differently, e.g.\ by using different technical implementations of the inverse-model's classifier.

\subsection{Decomposition of the loss}
\label{sec:decomposition}
In the following we show how the decomposition of the loss function $\mathcal{L}$ into the two components $\mathcal{L}_\mathrm{id}$ and $\mathcal{L}_\mathrm{manip}$ emerges.
Starting from the definition of the loss we can expand on the definition of the Kullback-Leibler divergence using the graphical model introduced in Fig.~\mref{fig:actions}B of the main manuscript:
\begin{align}
	\mathcal{L} =& D_\mathrm{KL}(p_\mathrm{EC}(a|x,x');q_\theta(a|x,x')) \nonumber \\
	=& \sum_{s=0}^1 ~ \sum_{b\in \mathcal{A}_\mathrm{manip}} p_\mathrm{EC}(a_\mathrm{id} = s|x,x') p_\mathrm{EC}(a_\mathrm{manip} = b|a_\mathrm{id}=s,x,x') \\
	& \quad \cdot \log \frac{p_\mathrm{EC}(a_\mathrm{id}=s|x,x') p_\mathrm{EC}(a_\mathrm{manip} = b|a_\mathrm{id}=s,x,x')}{q_\theta(a_\mathrm{id}=s|x,x') q_\theta(a_\mathrm{manip}=b|a_\mathrm{id}=s,x,x')}~,
\end{align}
where we have introduced $\mathcal{A}_\mathrm{manip}$ to accommodate all possibilities that $a_\mathrm{manip}$ can realize.

This expression can be grouped differently in order to simplify:
\begin{align}
	=& \sum_{s=0}^1 p_\mathrm{EC}(a_\mathrm{id}=s|x,x') \underbrace{\sum_{b \in \mathcal{A}_\mathrm{manip}}p_\mathrm{EC}(a_\mathrm{manip}=b|a_\mathrm{id}=s,x,x')}_{=1} \log \frac{p_\mathrm{EC}(a_\mathrm{id}=s|x,x')}{q_\theta(a_\mathrm{id}=s|x,x') } \nonumber\\
	&+ \sum_{s=0}^1 p_\mathrm{EC}(a_\mathrm{id}=s|x,x') \sum_{b \in \mathcal{A}_\mathrm{manip}}p_\mathrm{EC}(a_\mathrm{manip}=b|a_\mathrm{id}=s,x,x') \\
	& \quad \cdot \log \frac{p_\mathrm{EC}(a_\mathrm{manip}=b|a_\mathrm{id}=s,x,x')}{q_\theta(a_\mathrm{manip}=b|a_\mathrm{id}=s,x,x') }~,
\end{align}
which eventually gives rise to two separate Kullback-Leibler divergences:
\begin{align}
	=& D_\mathrm{KL}(p_\mathrm{EC}(a_\mathrm{id}|x,x'); q_\theta(a_\mathrm{id}|x,x')) \nonumber\\
	&+ \sum_{s=0}^1 p_\mathrm{EC}(a_\mathrm{id}=s|x,x') D_\mathrm{KL}(p_\mathrm{EC}(a_\mathrm{manip}|a_\mathrm{id}=s,x,x'); q_\theta(a_\mathrm{manip}|a_\mathrm{id}=s,x,x'))~.
\end{align}
Since $p_\mathrm{EC}(a_\mathrm{manip}|a_\mathrm{id}=1,x,x')$ and $q_\theta(a_\mathrm{manip}|a_\mathrm{id}=1,x,x')$ are fixed and 1 for the same, formally introduced, unknown $a_\mathrm{manip}$, we obtain:
\begin{align}
	\mathcal{L} =& D_\mathrm{KL}(p_\mathrm{EC}(a_\mathrm{id}|x,x'); q_\theta(a_\mathrm{id}|x,x')) \nonumber \\
	&+ p_\mathrm{EC}(a_\mathrm{id}=0|x,x') D_\mathrm{KL}(p_\mathrm{EC}(a_\mathrm{manip}|a_\mathrm{id}=0,x,x'); q_\theta(a_\mathrm{manip}|a_\mathrm{id}=0,x,x')) \nonumber\\
	=&: \mathcal{L}_\mathrm{id} + \mathcal{L}_\mathrm{manip}~.
\end{align}
\subsection{Instance discrimination as an upper bound to the identity-related inverse model loss}
\label{sec:upper_bound}
In the following, we provide proof that instance discrimination is an upper bound to $\mathcal{L}_\mathrm{id}$ as introduced in Section~\mref{sec:formalism} of the main manuscript.
Overall, the idea to achieve this, is to recognize that training the identity-related inverse model to always identify the correct positive example (i.e.\ related through $a_\mathrm{id}=0$) will at the same time allow this model to predict the case $a_\mathrm{id}=1$.

We begin by the definition of the loss for learning the identity-related inverse model:
\begin{align}
	\mathcal{L}_\mathrm{id} =& D_\mathrm{KL} ( p_\mathrm{EC}(a_\mathrm{id}|x,x');q_\theta(a_\mathrm{id}| x,x') ) \nonumber \\
	=& \,\mathrm{const} - \sum_{s=0}^1 p_\mathrm{EC}(a_\mathrm{id}=s|x,x') \log q_\theta(a_\mathrm{id}=s|x,x')~.
\end{align}
Since there are only two possibilities for $a_\mathrm{id} \in \{0, 1\}$ it follows that:
\begin{align}
	=& \,\mathrm{const} -p_\mathrm{EC}(a_\mathrm{id}=0|x,x')\log q_\theta(a_\mathrm{id}=0|x,x') \nonumber \\
	&- p_\mathrm{EC}(a_\mathrm{id}=1|x,x') \log (1 - q_\theta(a_\mathrm{id}=0|x,x'))~.\label{eq:expanded-loss-id}
\end{align}
From the definition of $q_\theta(a_\mathrm{id}=0|x,x')$ in Eq.~(\mref{eq:sw}) in the main manuscript, we have that $\sum_{x_n \in C}  q_\theta(a_\mathrm{id}=0|x,x_n) = 1 - q_\theta(a_\mathrm{id}=0|x,x')$, which further implies that we can take any $x'' \in C$ and obtain the inequality $ q_\theta(a_\mathrm{id}=0|x,x'') \leq 1 -  q_\theta(a_\mathrm{id}=0|x,x')$.


Inserting this inequality into Eq.~\eqref{eq:expanded-loss-id}, and due to the monotony of the logarithm, we obtain:
\begin{align}
	\mathcal{L}_\mathrm{id} \leq\,& \mathrm{const} -p_\mathrm{EC}(a_\mathrm{id}=0|x,x')\log q_\theta(a_\mathrm{id}=0|x,x') \nonumber \\
	&- p_\mathrm{EC}(a_\mathrm{id}=1|x,x') \log (q_\theta(a_\mathrm{id}=0|x,x''))~,
\end{align}

Finally, if on the other hand $p_\mathrm{EC}$ represents a perfect copy of $a_\mathrm{id}$, then we can define $x''$ to always represent the example that forms a positive pair with $x'$ (i.e.\ through $a_\mathrm{id} =0$) and obtain Eq.~(\mref{eq:objective}) in the main manuscript, which is the instance discrimination task of~\citep{chen2020simple}:
\begin{align}
	\mathcal{L}_\mathrm{id} \leq\,& \mathrm{const} - \log (a_\mathrm{id}=0|x,x'')~,
\end{align}
where $\mathrm{const}=0$ is a result of 0 entropy in $p_\mathrm{EC}(a_\mathrm{id}|x,x')$.

\section{Additional results}
\label{sec:additional}
\subsection{Visual depiction of the distribution of accuracies}

\begin{figure}[h]
	\centering
	\includegraphics[width=13.9cm]{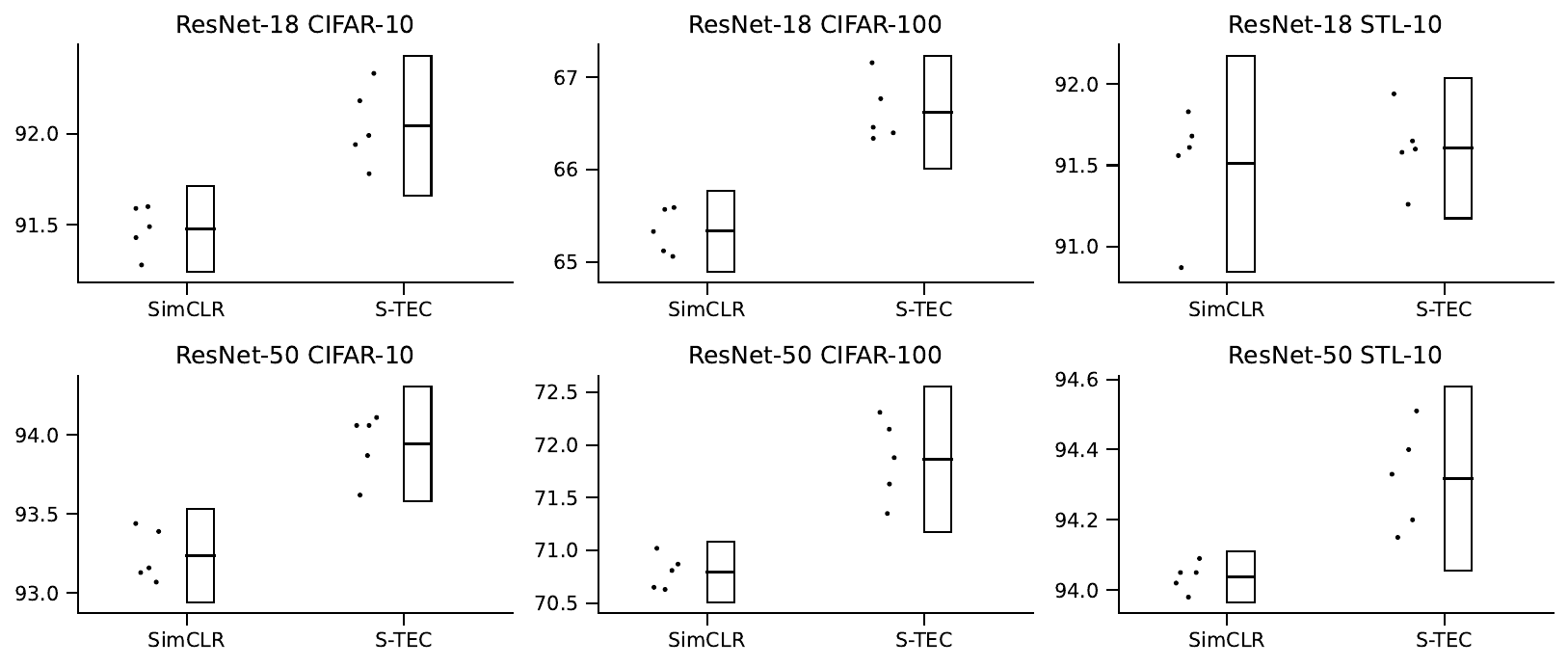}
	\caption{Visualization of performance distribution for SimCLR and S-TEC. In each setting 5 independent runs were conducted, resulting in different performances (points). These data points were used to obtain an estimated 95\% confidence interval (bar).}
	\label{fig:errorbars}
\end{figure}
In addition to the results presented in Table~\mref{tab:results} to~\mref{tab:imagenet} in the main manuscript, Fig.~\ref{fig:errorbars} depicts the distribution of performance values for SimCLR and S-TEC visually.

\subsection{Ablation}
We also performed an ablation study for ResNet-18s trained on CIFAR-100 to more specifically assess which components helped to improve the representations, as measured by linear classification accuracy, the most.
Subject to this study were two key mechanisms that were previously introduced: \textbf{categorical} and \textbf{egocentric} action representation.

\paragraph{Categorical action representation.}
The modelling of actions in a categorical manner induces a softmax cross entropy loss for the optimization of the manipulation-related inverse model, which we refer to as ``Classification''.
Alternatively, actions and the predictions thereof can be represented in their continuous form, which gives rise to a standard L2 regression loss, which we denote as ``L2 Regression'', pointing out that this strategy was employed by~\citet{lee2021improving}.

\paragraph{Egocentric action representation.}
On the other hand, our manipulation-related inverse model was trained to predict the actions that would be required to move from one view into the other based on its own perspective. This egocentric viewpoint is in contrast to the allocentric approach chosen in~\citep{lee2021improving}, where differences in the view are predicted based on the original image: i.e.\ for random cropping, differences in the cropping scale and differences of the crop's borders from the top and the left of the original image are predicted.

We tested the possible combinations of the choices for action representation (optimizing separately {$\lambda_\mathrm{manip} \in \{0.1, 0.2, 0.5, 1.0, 2.0\}$}) and report the average performance of 5 independent runs each in Fig.~\mref{fig:hp}A of the main manuscript.
These results confirm that categorical and egocentric action representations perform best as evaluated on linear classification accuracy.

\FloatBarrier
\subsection{Optimization progress}
\label{sec:optimization_progress}
\begin{figure}[h]
	\centering
	\includegraphics[width=13.9cm]{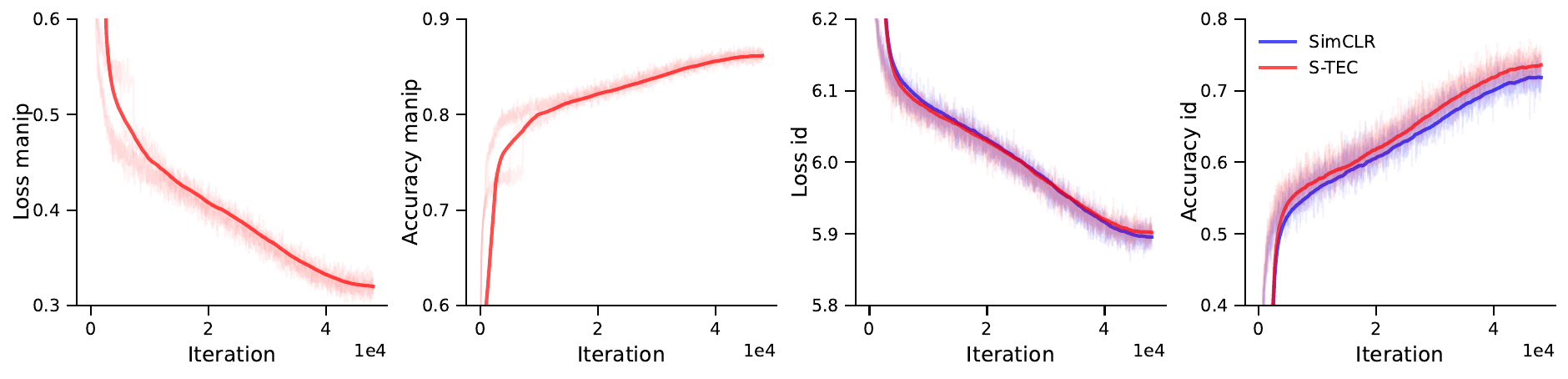}
	\caption{\textbf{ResNet-18 on CIFAR-10}: Progression of loss functions corresponding to the manipulation- and identity-related inverse model along with the accuracy of the respective task (training-set).}
	\label{fig:resnet18_cifar10_loss_curve}
\end{figure}

\begin{figure}[h]
	\centering
	\includegraphics[width=13.9cm]{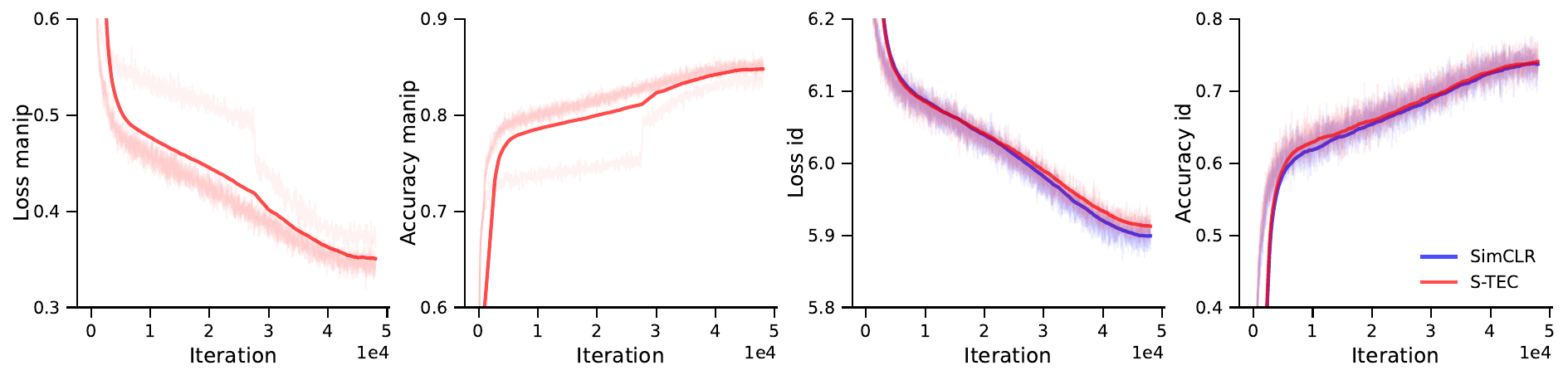}
	\caption{\textbf{ResNet-18 on CIFAR-100}: Progression of loss functions corresponding to the manipulation- and identity-related inverse model along with the accuracy of the respective task (training-set).}
	\label{fig:resnet18_cifar100_loss_curve}
\end{figure}

\begin{figure}[h]
	\centering
	\includegraphics[width=13.9cm]{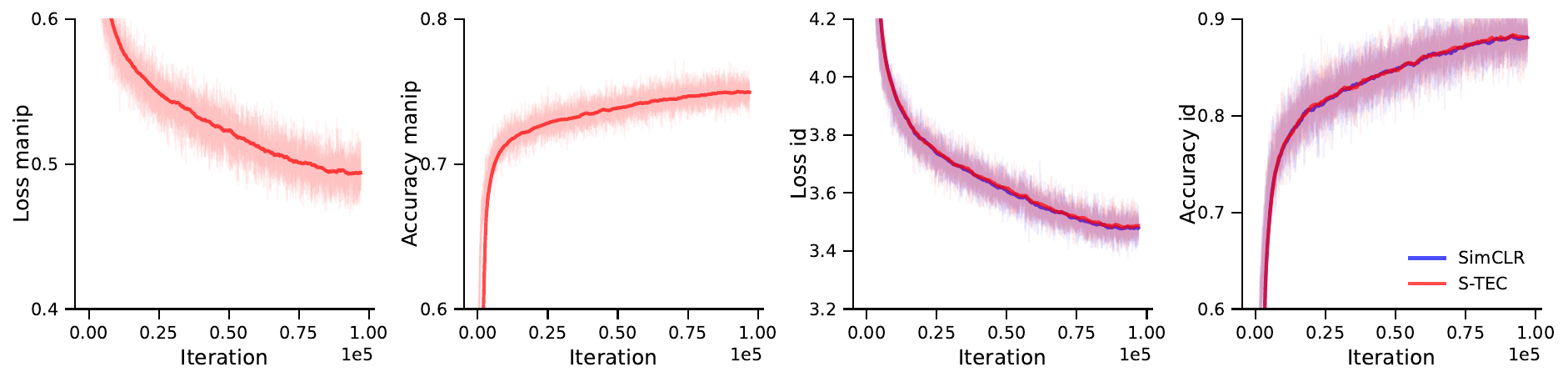}
	\caption{\textbf{ResNet-18 on STL-10}: Progression of loss functions corresponding to the manipulation- and identity-related inverse model along with the accuracy of the respective task (training-set).}
	\label{fig:resnet18_stl10_loss_curve}
\end{figure}

\begin{figure}[h]
	\centering
	\includegraphics[width=13.9cm]{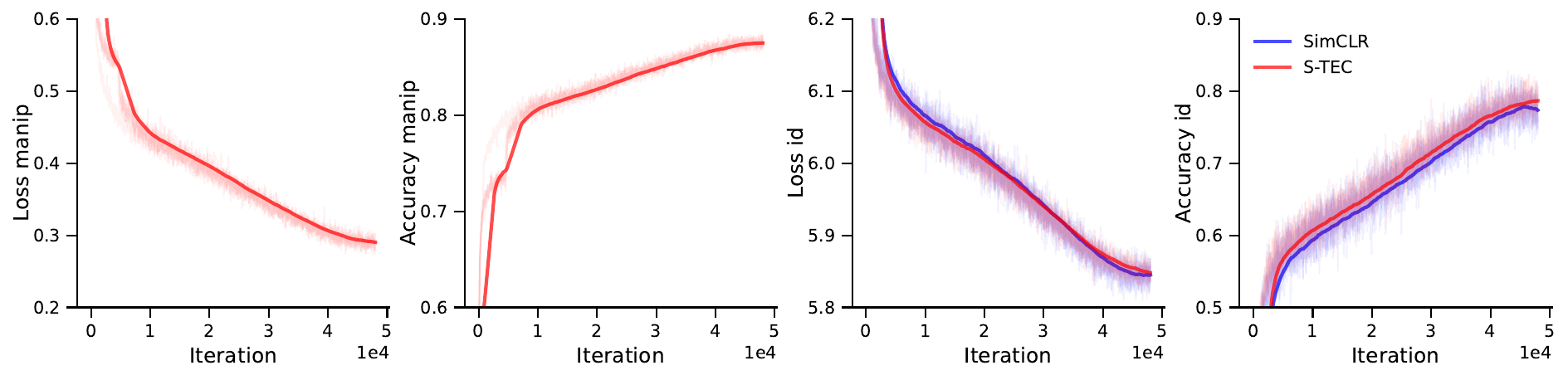}
	\caption{\textbf{ResNet-50 on CIFAR-10}: Progression of loss functions corresponding to the manipulation- and identity-related inverse model along with the accuracy of the respective task (training-set).}
	\label{fig:resnet50_cifar10_loss_curve}
\end{figure}

\begin{figure}[h]
	\centering
	\includegraphics[width=13.9cm]{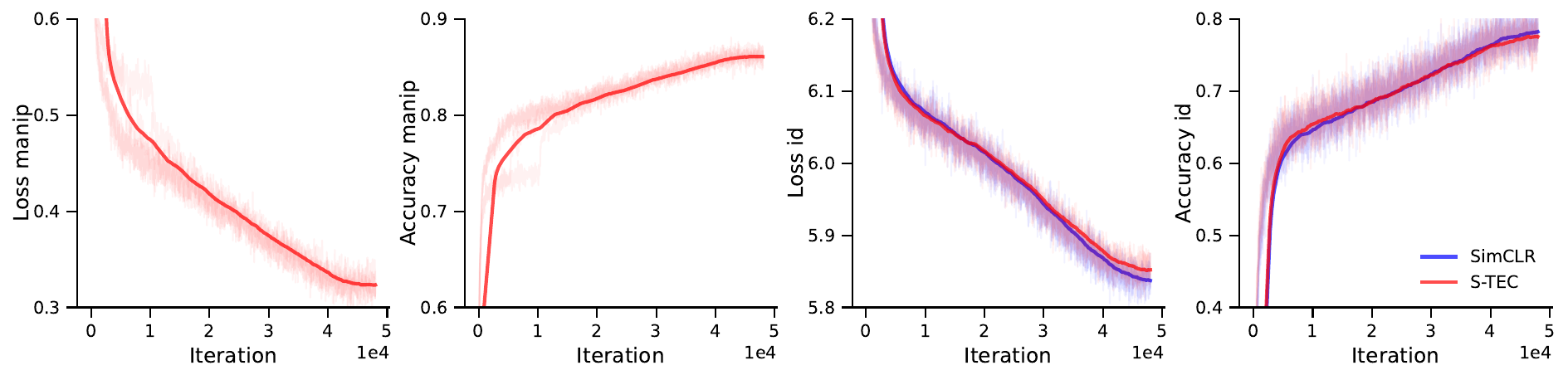}
	\caption{\textbf{ResNet-50 on CIFAR-100}: Progression of loss functions corresponding to the manipulation- and identity-related inverse model along with the accuracy of the respective task (training-set).}
	\label{fig:resnet50_cifar100_loss_curve}
\end{figure}

\begin{figure}[h]
	\centering
	\includegraphics[width=13.9cm]{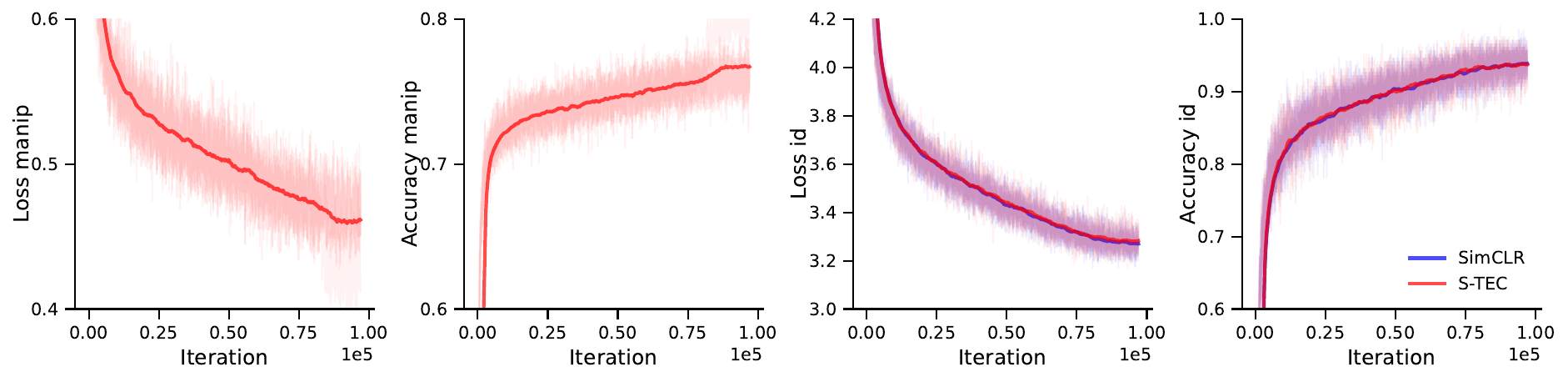}
	\caption{\textbf{ResNet-50 on STL-10}: Progression of loss functions corresponding to the manipulation- and identity-related inverse model along with the accuracy of the respective task (training-set).}
	\label{fig:resnet50_stl10_loss_curve}
\end{figure}

For insight in the optimization dynamics, we provide learning curves for runs of SimCLR and S-TEC on the datasets of CIFAR-10 (see Fig.~\ref{fig:resnet18_cifar10_loss_curve} and~\ref{fig:resnet50_cifar10_loss_curve}), CIFAR-100 (see Fig.~\ref{fig:resnet18_cifar100_loss_curve} and~\ref{fig:resnet50_cifar100_loss_curve}) as well as loss curves in the case of STL-10 (see Fig.~\ref{fig:resnet18_stl10_loss_curve} and~\ref{fig:resnet50_stl10_loss_curve}).
All of the provided curves were obtained using 5 independent runs for each scenario that was considered.
We report the averages of these as bold curves, which were additionally processes using a moving average filter.
Unprocessed individual metrics are shown as thin transparent lines.

We report in each scenario the following metrics:
\begin{enumerate}
	\item Loss of the manipulation-related objective (only for S-TEC),
	\item accuracy of the manipulation-related inverse model (only for S-TEC), which is defined as the average accuracy that this inverse models picks the correct action clusters (measured on the training set),
	\item loss of the identity-related inverse model, and
	\item accuracy of the identity-related inverse model, which is reported as the fraction of positive views $x$ and $x''$ being correctly identified, see also~\ref{sec:upper_bound}.
\end{enumerate}

